\documentclass[10pt,journal]{IEEEtran}

\usepackage{cite}

\ifCLASSINFOpdf
\usepackage[pdftex]{graphicx}
\else
\fi
\usepackage{ifpdf}
\usepackage{times}
\usepackage{epsfig}
\usepackage{graphicx}
\usepackage{amsmath}

\usepackage{amssymb}
\usepackage{color}
\usepackage{subfig}
\usepackage{tabu}
\usepackage{tabularx}
\usepackage{multirow}
\usepackage[abs]{overpic}
\usepackage{algorithm}
\usepackage{algorithmic}
\usepackage{wrapfig}
\usepackage{array}
\usepackage{stfloats}
\usepackage{booktabs}
\usepackage[table,xcdraw]{xcolor}
\usepackage{geometry}
\graphicspath{{./Imgs/}{C:/Users/sse307/Desktop/latexv1/Imgs/}}    
\DeclareGraphicsExtensions{.pdf,.png,.jpg}
\newcolumntype{C}[1]{>{\centering\arraybackslash}p{#1}}

\newcommand{\figref}[1]{Fig.~\ref{#1}}
\newcommand{\tabref}[1]{Table.~\ref{#1}}

\newcommand{\equref}[1]{Eq. (\ref{#1})}

\def\etal{{\em et al.~}}

\begin{document}
%
\title{Salient Instance Segmentation via Subitizing and Clustering}
%
%
%

\author{Jialun Pei, He Tang, Chao Liu, and Chuanbo Chen
  \thanks{* He Tang is the corresponding author.}
  \thanks{Jialun Pei is with the School of Software Engineering and School of Computer Science and Technology, Huazhong University of Science and Technology, 1037 Luoyu Road, Wuhan, 430074, China (e-mail: peijl@hust.edu.cn)}
  \thanks{He Tang, Chao Liu, and Chuanbo Chen are with the School of Software Engineering, Huazhong University of Science and Technology, 1037 Luoyu Road, Wuhan, 430074, China (e-mail: hetang@hust.edu.cn; chuanboc@163.com; hanshanlc@hust.edu.cn)}
}

\maketitle

\begin{abstract}
  The goal of salient region detection is to identify the regions of an image that attract the most attention. Many methods have achieved state-of-the-art performance levels on this task. Recently, salient instance segmentation has become an even more challenging task than traditional salient region detection; however, few of the existing methods have concentrated on this underexplored problem. Unlike the existing methods, which usually employ object proposals to roughly count and locate object instances, our method applies salient objects subitizing to predict an accurate number of instances for salient instance segmentation. In this paper, we propose a multitask densely connected neural network (MDNN) to segment salient instances in an image. In contrast to existing approaches, our framework is proposal-free and category-independent. The MDNN contains two parallel branches: the first is a densely connected subitizing network (DSN) used for subitizing prediction; the second is a densely connected fully convolutional network (DFCN) used for salient region detection. The MDNN simultaneously outputs saliency maps and salient object subitizing. Then, an adaptive deep feature-based spectral clustering operation segments the salient regions into instances based on the subitizing and saliency maps. The experimental results on both salient region detection and salient instance segmentation datasets demonstrate the satisfactory performance of our framework. Notably, its AP$^{r}$@0.7 reaches 60.14\% in the salient instance dataset, surpasses the state-of-the-art methods by about 5\%.

\end{abstract}


%
\IEEEpeerreviewmaketitle

\section{Introduction}
%
%
%
%
\IEEEPARstart{T}{he} seminal work by Itti \etal illustrates that the most arresting objects are usually visually salient \cite{itti1998model}. Salient object detection models  aim at highlighting the most attractive regions of an image, simulating the visual attention process in human visual systems \cite{lei2016universal}. In contrast to image segmentation and target recognition, saliency maps are determined by only the most salient objects \cite{borji2014salient}. Salient object detection has recently attracted considerable interest in computer vision tasks such as scene understanding \cite{cheng2010repfinder}, image editing \cite{wu2014hierarchical}, image retrieval \cite{gao2015database}, video summarization \cite{hua2005generic} and robotic perception \cite{sugano2010calibration}.

Over the past two decades, saliency detection has played an increasingly important role in computer vision problems. The accuracy of salient object detection has improved rapidly due to the renaissance of convolutional neural network (CNN) models \cite{zhao2015saliency, li2015visual}, which have shown superior performance over traditional solutions \cite{huo2017iterative, ye2017salient, quan2017unsupervised}. Due to the multilevel and multiscale features extracted by CNNs, the most salient objects can be captured with high precision \cite{xiao2018deep}. However, the previous works like \cite{zhang2017amulet, liu2016dhsnet, zhang2018bi} focused only on the salient regions and overlooked the individual instances of salient regions. Currently, a new challenging task has gradually attracted widespread attention: instance-level salient object segmentation (or salient instance segmentation for short), which was first proposed by \etal \cite{li2017instance}. Compared to salient object detection, salient instance segmentation attempts to discriminate individual object instances in the salient regions of an image or scene. The resulting instance-level saliency maps foster a more detailed analysis by labeling each instance with an accurate pixelwise segmentation map. Compared to general instance segmentation, salient instance segmentation predicts salient object instances without categorizing the objects and has more specific and in-depth application areas than does salient object detection, including target recognition, driver assistance and image captioning \cite{karpathy2015deep}.

Salient instance segmentation is challenging because the salient regions exhibit boundaries that are often obscured or overlapped with other instances that possess similar features. Moreover, the high occlusion probability and the diverse shape deformations have a negative effect on the ability to distinguish intersecting instances \cite{liang2018proposal}. In general, the existing methods locate individual instances in salient regions with the aid of proposals. Li \etal combined salient object proposals and contours to generate the final instances \cite{li2017instance}.  S4Net also uses RPN (Region Proposal Network) to generate a number of proposals, and then uses SID (salient instance discriminator) to segment the salient instances \cite{fan2019s4net}. However, the proposal-based approach has some important drawbacks. First, region proposal schemes typically generate hundreds of candidate bounding boxes, and filtering these numerous proposals to obtain an optimized subset is inefficient \cite{cheng2014bing, uijlings2013selective}. Second, the proposed bounding boxes do not produce boundaries for strongly occluded objects. Consequently, the post-processing algorithm has considerably difficulty detecting highly accurate object boundaries.

\begin{figure}[!t]
  \centering
  \includegraphics[width=\linewidth]{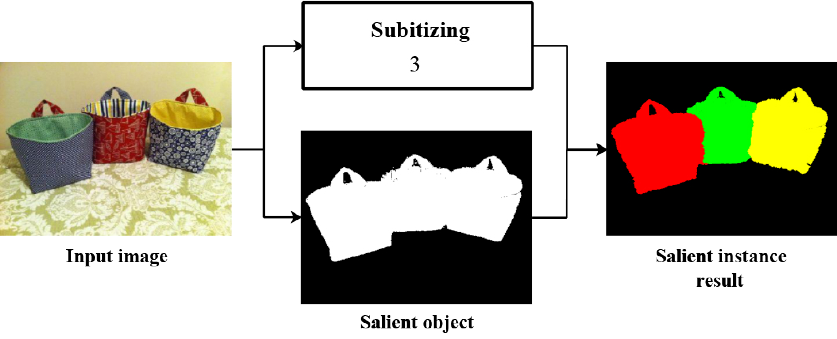} 
  \caption{The pipeline of the proposed method illustrates the salient instance segmentation.}
  \label{pipelinemethod}
  \vspace{-0.1in}
\end{figure}


In this paper, we propose a proposal-free method to effectively solve the salient instance segmentation task. The pipeline of the salient instance segmentation process used by the proposed method is illustrated in \figref{pipelinemethod}. Different from existing methods based on object proposals, the proposal-free method applies salient object subitizing process rather than proposals to avoid the aforementioned disadvantages of proposals \cite{zhang2017salient}. In addition, some proposal methods used for object detection are category-dependent, which makes them inappropriate for use in salient instance segmentation because the goal is to distinguish individual instances rather than perform category-level segmentation. We focus on category-independent salient instance segmentation and propose a multitask neural network that applies two parallel branches a densely connected subitizing network (DSN) to predict the number of instances and a densely connected fully convolutional network (DFCN) for salient region detection. Then, an adaptive deep feature-based spectral clustering operation combines the subitizing and saliency maps to generate promising instance-level salient object segmentation results \cite{ng2002spectral}. The proposed salient instance segmentation task is divided into three subtasks: saliency detection, subitizing prediction, and clustering of the salient instances.

First, we use a DSN to perform salient object subitizing. We were motivated to use a subitizing model to segment salient instances by Zhang \etal \cite{zhang2017salient}. The fully connected layer of DenseNet is applied to predict the number of instances, and the accuracy rate of this approach is better than that of GoogleNet, which was applied in \cite{zhang2017salient}. Second, we extend DenseNet \cite{huang2017densely} to our DFCN by adding upsampling layers. It is worth noting that a fully convolutional network (FCN) is regarded as a CNN extension that recovers the full resolution of the input images, while CNNs are appropriate for image classification prediction tasks \cite{long2015fully}. To reduce the resolution loss and retain the multiscale features from the downsampling path, we build skip connections between the downsampling and upsampling paths. In addition, a fully connected Conditional Random Field (CRF) works as a pixelwise refined model to purify the salient region maps \cite{li2016deep}. Our network outputs full-resolution salient maps with high accuracy and outperforms other state-of-the-art methods. We provide salient region detection result comparisons in Section \uppercase\expandafter{\romannumeral4}. Third, the subitizing operation provides support for the initial number of clusters $k$ of the spectral clustering algorithm. Therefore, spectral clustering can be used to merge the salient region maps and perform subitizing to output the final results. To improve the spectral clustering effectiveness, the deep features of DFCN are extracted as input to the spectral clustering algorithm. In addition, we apply the simple linear iterative clustering (SLIC) algorithm to the clustering process, which boosts the instance edge segmentation effect during the k-means spectral clustering step\cite{achanta2012slic}. At the clustering step in spectral clustering, we adopt a k-means algorithm that improves the selection of initial clustering centers. The proposed MDNN achieves excellent performance and surpasses all the existing methods when applied to the only available salient instance segmentation dataset (dataset1K) in \cite{li2017instance}. Moreover, our saliency model outperforms other state-of-the-art models on public salient object detection datasets. In summary, our contributions are as follows:
\begin{itemize}
  \item We propose a multitask densely connected neural network (MDNN) to address the challenging salient instance segmentation task. The proposed method is proposal-free and category-independent, which improve the robustness of the algorithm.
  \item A subitizing process is adopted to predict the number of instances using a DSN instead of region proposal methods. Moreover, we detect salient objects by upgrading DenseNet to a DFCN that produces accurate, full-resolution saliency maps.
  \item We apply an adaptive spectral clustering operation to produce the final salient instances. We improve the adjacency matrix $W$ of spectral clustering by adding multiscale deep features, which improves the resulting performance. In addition, we develop an adaptive quantile strategy instead of using random selection that allows us to flexibly and automatically select the initial clustering centers.
\end{itemize}

The remainder of this paper is organized as follows. Section \uppercase\expandafter{\romannumeral2} introduces the related works. Section \uppercase\expandafter{\romannumeral3} describes the theoretical basis and the architecture of the proposed method. Section \uppercase\expandafter{\romannumeral4} presents the experimental comparisons and discussions. Finally, Section \uppercase\expandafter{\romannumeral5} concludes the paper.

\section{RELATED WORK}

Thanks to the deep convolutional neural networks that have been developed for salient object detection, the current salient object segmentation method is good enough to function as an intermediate process for salient instance segmentation \cite{huang2015salicon}. However, due to the category-independent nature of salient object segmentation, salient instance segmentation cannot be regarded as simply a subbranch of instance segmentation. To understand the task of salient instance segmentation, in essence, we can consider salient object detection, instance-level semantic segmentation and salient instance segmentation as three different types of image segmentation problems.

\subsection{Salient object detection}

Salient object detection is usually divided into two phases: the first phase detects the most salient regions in the image, and the second phase accurately segments salient objects from the salient regions \cite{borji2013state}. Liu \etal \cite{liu2011learning} and Achanta \etal \cite{achanta2009frequency} presented salient region detection as an outgrowth of the binary object segmentation problem. Traditional salient object detection depends primarily on traditional machine learning methods such as bottom-up and top-down methods based on multilevel features \cite{goferman2012context} to build saliency models. For instance, Cheng \etal extracted salient regions by computing the global contrast in the image \cite{cheng2015global}; Perazzi \etal considered salient object detection as a filtering problem \cite{perazzi2012saliency}; Zhang \etal achieved high performance results based on a Boolean map approach \cite{zhang2013saliency}. With the development of deep CNNs, salient object detection has gradually become dominated by deep neural network model methods. Kruthiventi \etal used a fully convolutional neural network to simultaneously perform eye-fixation and salient object detection \cite{kruthiventi2016saliency}. Li \etal introduced semantic features into a multitask convolutional network to assist in salient object detection \cite{li2016deepsaliency}. These latest end-to-end networks improve both accuracy and efficiency compared to the earlier traditional methods. Using newer neural network models, the salient object detection problem can be determined using a fully convolutional neural network similar to semantic segmentation. The proposed multitask densely connected neural network (MDNN) sets the output layer of the DFCN to two channels, which respectively classify the salient regions and background in the image.

\begin{figure*}[!t]
    \centering
    \includegraphics[width=\linewidth]{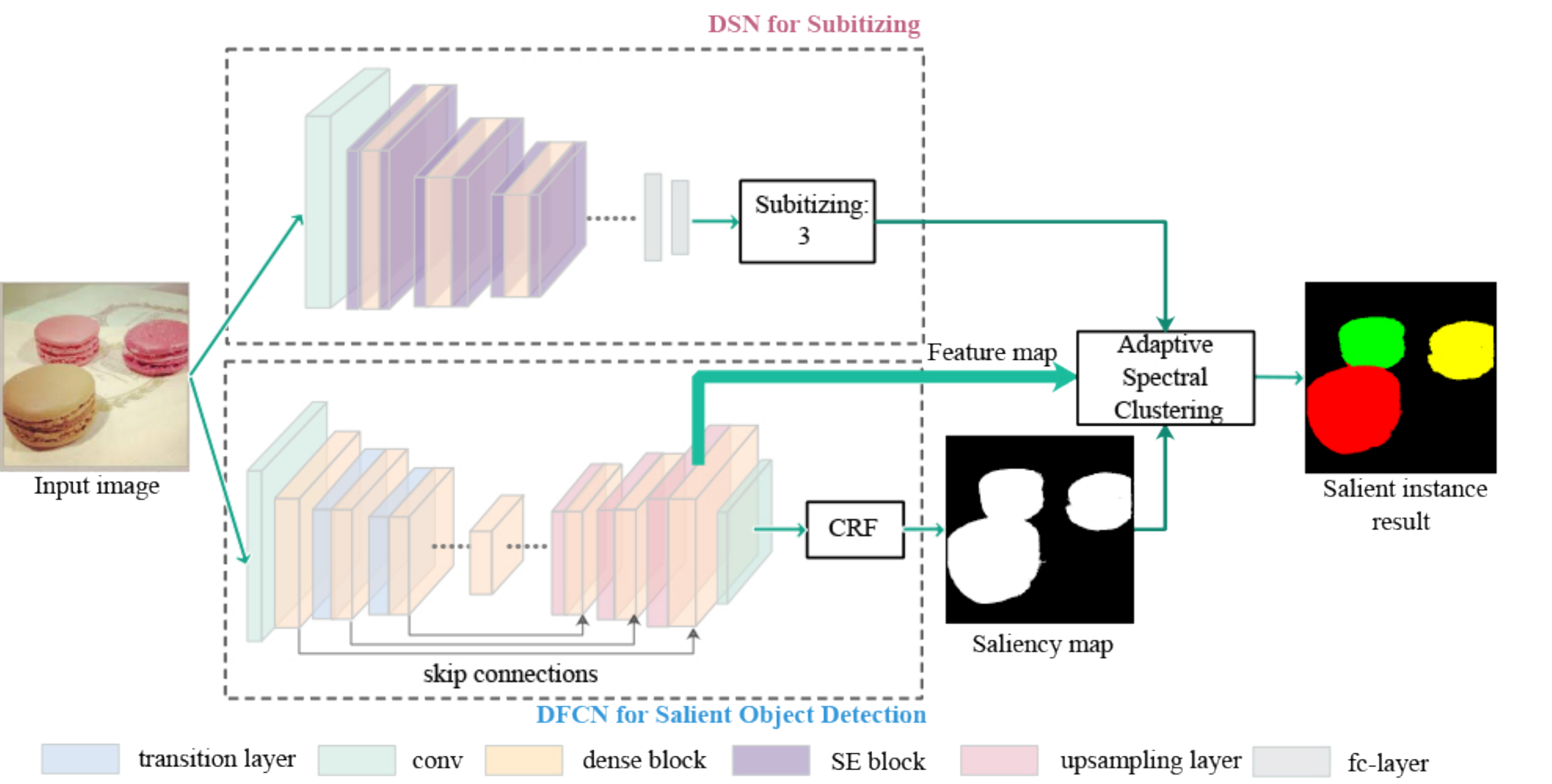} 
    \caption{The proposed overall MDNN framework for salient instance segmentation. The upper dashed box is the subitizing network for predicting number of instances. The bottom dotted box contains the DFCN and the fully connected CRF for producing saliency maps. The skip connections are used to reuse hierarchical features in the upsampling path. The feature from the last dense block layers are extracted for spectral clustering.}
    \label{framework}
\end{figure*}

\subsection{Instance-Aware semantic segmentation}

The earlier instance-aware semantic segmentation is defined as a multitask operation consisting of object detection and semantic segmentation. The limitation of this task is that object detection estimates the bounding boxes only by using segmentation methods, and semantic segmentation simply predicts the category of each pixel. To overcome this constraint, Hariharan \etal presented simultaneous detection and segmentation (SDS) to detect all instances of a category in an image \cite{hariharan2014simultaneous}. Subsequently, Dai \etal used a multitask cascade network to detect the proposals and then extracted contours to classify the objects \cite{dai2016instance}. In \cite{dai2016instancee}, the authors proposed a model named instance-sensitive fully convolutional networks that integrate feature proposals in different locations into a complete mask, which is different from traditional FCNs. In particular, a Mask R-CNN framework was presented to promote the Faster R-CNN architecture by adding a branch to predict an object mask \cite{he2017mask, girshick2015fast}. Despite the relatively fast development of semantic instance segmentation, these frameworks rely on defined categories. In contrast, for category-independent salient instance segmentation, it is unnecessary to predefine classes before segmentation.

\subsection{Salient instance segmentation}

Salient object detection should must be able to solve two problems: they must be able to determine the number of objects and detect the location of each object, both of which are important aspects of salient instance segmentation. Zhang \etal utilized a CNN to generate salient object proposals and predicted the proposal locations with a subset optimization framework based on the maximum a posteriori principle \cite{zhang2016unconstrained}. This approach promoted the salient objects from region-level to instance-level; thus, each instance could correspond to a bounding box for the first time. Li \etal were the first to address the instance-level salient object segmentation task \cite{li2017instance}. The developed method used the multiscale refinement network (MSRNet) to produce the saliency maps and contour maps and then used subset optimization to refine the number of proposals. A fully connected CRF performed postprocessing to generate the final results. In addition, they compiled a new dataset for salient instance segmentation. However, their method depended heavily on the CRF process for edge detection, and the quality of the optimization proposals was affected by the subset optimization.Recently, Fan \etal proposed an end-to-end neural network to solve the problem of salient instance segmentation, called S4Net \cite{fan2019s4net}. The method first used a single-shot object detector to detect the approximate position of the instances, and proposed to use a ternary masking instead of the traditional binary masking, which is similar to the center-around difference theory in the saliency detection \cite{treisman1980feature}.

All the above-mentioned works depend on region proposals to locate and predict the number of instances. However, this approach is both inefficient and complicated due to the number of proposals generated and the need to execute postprocessing steps. To avoid this negative effect, we propose a proposal-free multitask NN framework to generate the salient instance segmentation results. Performing subitizing instead of the proposal method can boost the robustness of the final results.

\begin{figure*}[!t]
    \centering
    \includegraphics[width=\linewidth]{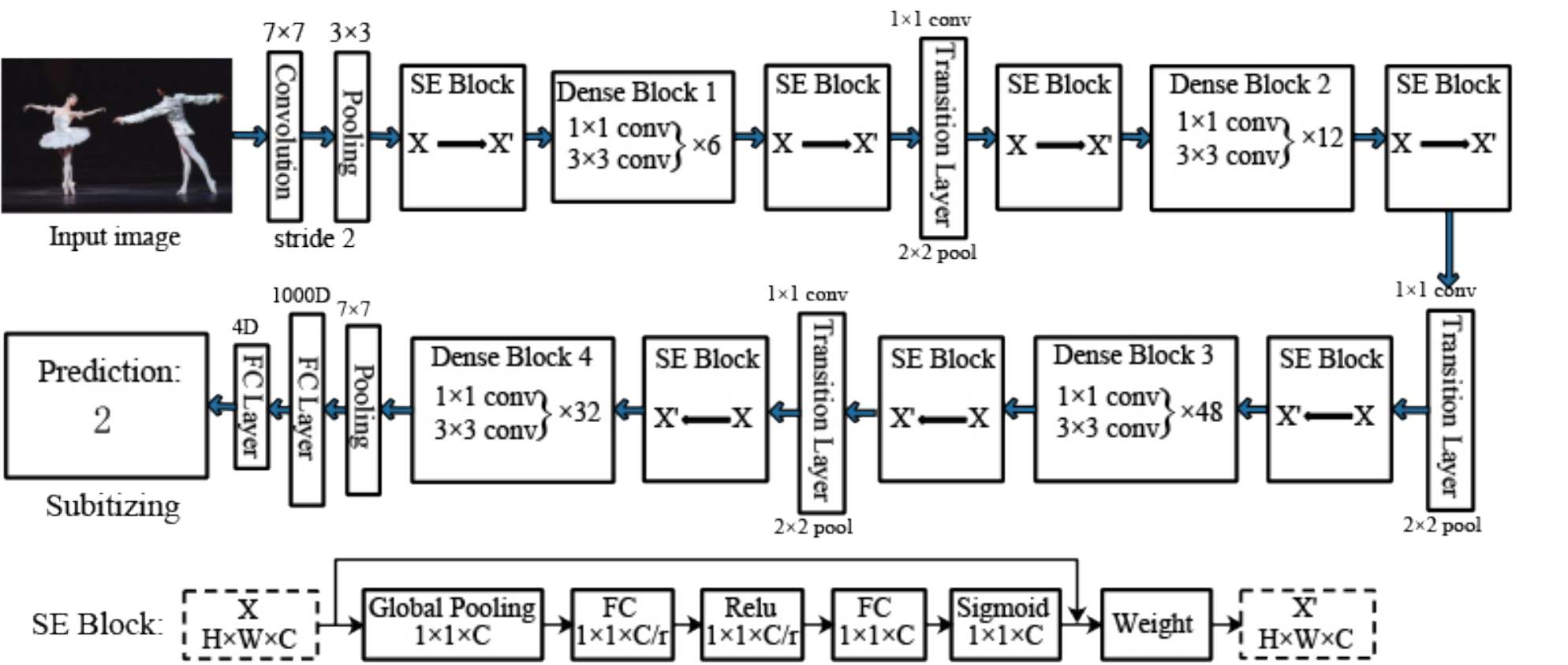} 
    \caption{Details of the DSN network in MDNN, the bottom of Fig. 3 is the schema of the SE block. $X$ and ${X}'$ represent the original feature and the recalibration of feature respectively. $r$ is the reduction ratio, which value is set to 16 in this paper.}
    \label{DSNnetwork}
    \vspace{-0.1in}
\end{figure*}

\section{PROPOSAL-FREE SALIENT INSTANCE SEGMENTATION}

\subsection{Overall framework}

\figref{framework} shows the overall architecture of the proposed proposal-free salient instance segmentation method. The framework consists of three main processes. (1) a densely connected subitizing network (DSN) that predicts the number of instances; (2) a densely connected fully convolutional network (DFCN) that performs salient region detection; and (3) adaptive spectral clustering using deep features to perform salient instance segmentation. Given an input image, we first use the DSN to predict the number of salient instances. The proposed DFCN is tailored to perform salient region detection. By extending the upsampling layers in the network and adding skip connections between the downsampling and upsampling paths the DFCN achieves a better pixelwise salient detection capability. A fully connected CRF \cite{krahenbuhl2011efficient} is adopted as a postprocessor to refine the salient region maps produced by the DFCN. Then, salient instances are segmented from the subitizing and salient regions based on adaptive spectral clustering.

\subsection{Densely connected subitizing network for instance-number prediction}

The main idea of the proposal-free method is to use salient object subitizing instead of an object region proposal method. Although object proposal methods locate the objects and estimate the number of instances by optimizing the number of bounding boxes, subitizing can directly predict the number of salient objects from the salient regions already located by the salient region detection task \cite{alexe2010object}. In other words, optimized object proposals may include nonsalient objects, while in contrast, subitizing directly predicts the salient number of objects. However, background distraction is usually the main challenge, which causes detection errors \cite{he2017delving}. Since Zhang \etal \cite{zhang2017salient} proposed salient object subitizing, the operation has been neither widely advanced nor widely applied thus far. In light of the concept of instance-level salient object segmentation, it is appropriate to use subitizing to accurately enumerate the number of instances.

\begin{figure*}[!t]
    \centering
    \includegraphics[width=\linewidth]{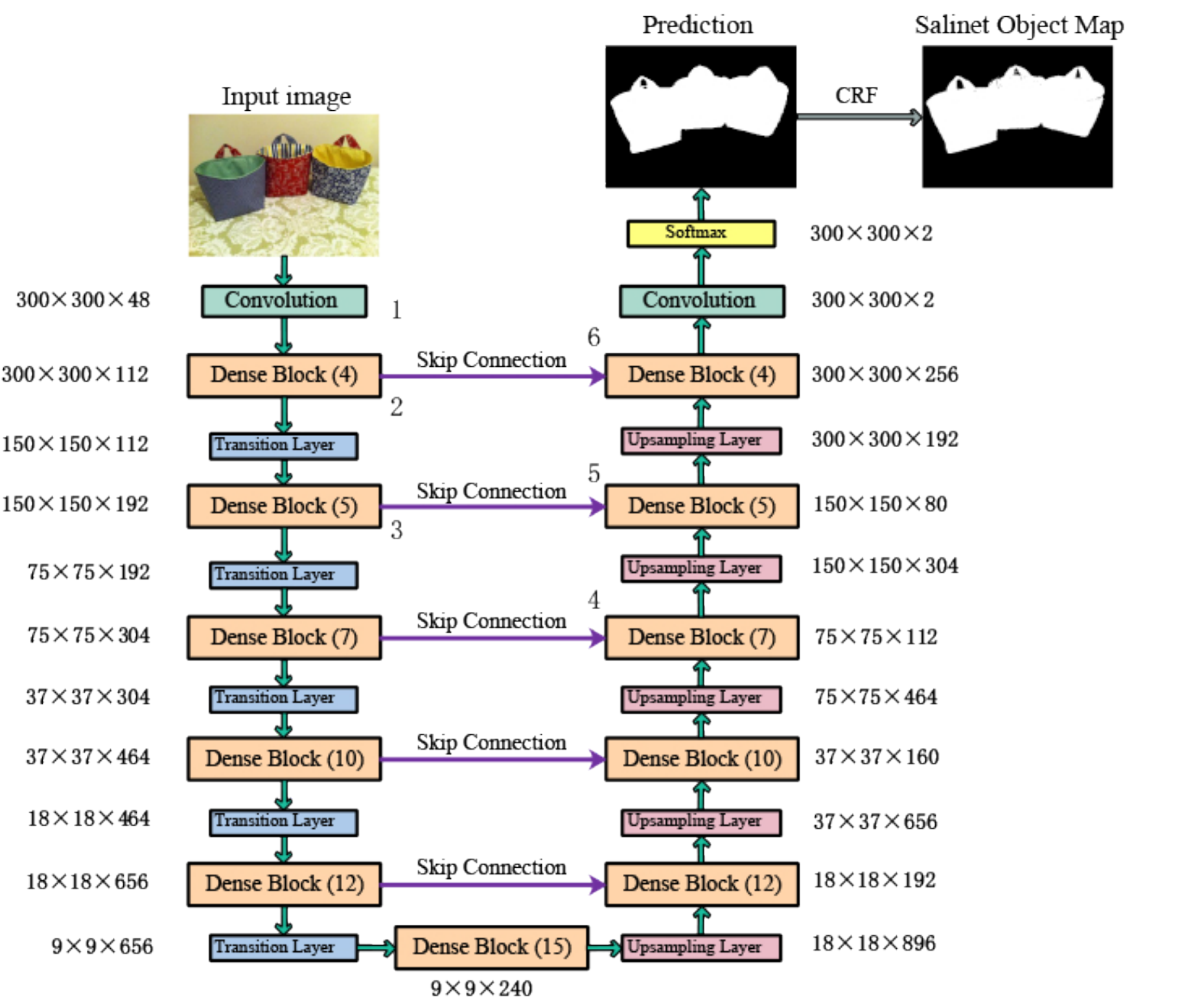}
    \caption{The architecture of the U-shaped DFCN. The numbers in the dense block boxes indicate the number of layers contained in each dense block. The number 1 through 6 indicate the indexes of the candidate feature maps of the proposed adaptive spectral clustering, which discussed in the ablation study of Section \uppercase\expandafter{\romannumeral4}.}
    \label{DFCNnetwork}
    \vspace{-0.1in}
\end{figure*}

As mentioned earlier, the purpose of the DSN in the proposed MDNN is to predict the number of instances in an input image. Different from the way AlexNet and GoogleNet in SOS predicted the number of salient objects \cite{krizhevsky2012imagenet}, we construct the DSN based on DenseNet. Compared with AlexNet and GoogleNet, DenseNet requires fewer parameters and preserves richer features based on how it performs concatenation \cite{huang2017densely}. More importantly, we embed the SE (Squeeze-and-Excitation) block for improving the representational power of the densely subitizing network by explicitly modelling the interdependencies between the channels of its convolutional features. The detailed DSN structure for subitizing is illustrated in \figref{DSNnetwork}. In our work, the subitizing task is viewed as a classification problem. First, we resize the input image to $224\times224$ to adapt the image to the downsampling process. The size of the initial convolution kernel is $7\times7$ with a stride of 2. Followed by a $3\times3$ max pooling layer, the first SE block is connected, which is illustrated in the bottom of Fig. 3. It is designed to improve the representational capacity of a network by enabling it to perform dynamic channel-wise feature recalibration. We consider the input features $X\in R^{H*W*C}$ as $X=[x_{1},x_{2},...,x_{c}]$ where $X_{i}\in R^{H*W}$is the $i\textbf{-}$th channel of $X$ and $C$ is the total channel number. Firstly, the global average  pooling is worked on each $x_{i}$ to obtain a channel-wise feature vector $v\in R^{C}$. Then, two fully connected (FC) layers are used to capture channel-wise dependencies. Like \cite{hu2018squeeze}, we encode the channel-wise feature vector by forming a bottleneck with two FC layers around the non-linearity to limit model complexity and aid generalization. After adjusting the channel weight by the sigmoid operation, the output features are adapted to the input-specific descriptor $v$. the Squeeze-and-Excitation process is described as:
\begin{equation}
    X^{'}=(\varphi (fc_{2}(\eta (fc_{1}(v,W_{1})),W_{2})))\cdot X ,
\end{equation}
where $\eta$ refers to the ReLU function, $\varphi$ refers to sigmoid operation, $fc$ refers to FC layers, $W_{1}\in R^{\frac{c}{r}\times c}$ and $W_{2}\in R^{c\times \frac{c}{r}}$. The parameter $W_{1}$ is used to reduce the dimension of the feature and $W_{2}$ is worked to recover the dimension of the feature for adapting the next layer. The reduction ratio $r$ in $W_{1}$ and $W_{2}$ is an important hyperparameter. According to \cite{hu2018squeeze}, $r$ is set to 16 for all experiments in this paper. We added a total of 7 SE blocks in the structure of DSN.

Followed by the first SE block, the feature $X^{’}$ is transmitted to the dense block which embrace 6 layers. Each layer in the dense block contains a $1\times1$ convolutional layer followed by a $2\times2$ average pooling layer. The main advantage of this DSN lies in how it performs concatenation manner; it connects each layer to every other layer in a feed-forward fashion \cite{huang2017densely}. Given that a dense block contains $L$ layers, we define a nonlinear transformation $H_l (\cdot)$ as a successive composite function. The function embraces three operations: batch normalization (BN) \cite{ioffe2015batch}, a rectified linear unit (ReLU) \cite{glorot2011deep} and a $3\times3$ convolution \cite{huang2017densely}. We assume that $x_{l}$ is the output of the $l\textbf{-}$th layer, which is defined as:
\begin{equation}
    x_l=H_l ([x_{0},x_{1},x_{2},...,x_{l-1}]),
\end{equation}
where $[x_{0},x_{1},x_{2},...,x_{l-1}]$ represents the series of connected feature maps. This type of connectivity pattern alleviates feature loss and encourages feature reuse. As shown in \figref{DSNnetwork}, the proposed network has four dense blocks, each of which embeds 6, 12, 48 and 32 layers, respectively. The feature map sizes produced in these four dense blocks are $56\times56$, $28\times28$, $14\times14$ and $7\times7$, respectively. In addition, we insert three transitional layers between two contiguous dense blocks composed of $1\times1$ convolution followed by $2\times2$ average pooling. At the end of the network, the classification layer consists of a $7\times7$ average pool, a 1000-D fully connected layer and a 4-D fully connected layer. Normally, the human visual system is able to consider only up to 4 salient instances simultaneously without thinking \cite{he2017delving}. Using the dataset from \cite{li2017instance}, we do not know how many salient instance images to identify, so we set the last FC layer to 4-D to predict 1, 2, 3 and 4+ salient instances in the image data.

\subsection{Densely connected fully convolutional network for salient region detection}

The goal of the salient region detection task is to label the salient and interesting regions from the image \cite{liu2011learning}. Because the final mission is to segment the salient instances, it is important for the salient detection procedure to produce precise salient regions. We proposed the densely connected fully convolutional network (DFCN) to detect salient regions; this network was inspired by DenseNet and FCNs \cite{long2015fully}.

The DFCN is described in \figref{DFCNnetwork}. The framework pipeline looks similar to a U-shape after adding the deconvolution process based on DenseNet \cite{ronneberger2015u}. The proposed DFCN takes advantage of the concatenation process in the dense blocks to build an FCN-like architecture for generating saliency maps of arbitrary sizes. To further preserve the downsampling layer features, we build bridges between the upsampling and downsampling layers called "skip connections" to ameliorate the resolution loss and boost the completeness of the feature at the upsampling step.

Given an input image, the size of the first convolution is $3\times3$; and 5 dense blocks are concatenated with the same number of transition layers. The transition layers are structured the same as those in DenseNet, but the number of layers in the dense blocks are different, with 4, 5, 7 10 and 12 layers, respectively. The bottom of the network embeds a 15-layer dense block followed by an upsampling layer. An upsampling operation consisting of a $3\times3$ transposed convolution with a stride of 2. The upsampling path is used not only to recover the input spatial resolutions but also to upsample the feature maps. The output of the upsampling layer is combined with the features from the skip connections to form the input of the next dense block. In the upsampling path, the dense block corresponds to the downsampling path, and the upsampling layer compensates for the pooling operation in the transition layers. The input to the last dense block consists of the summed information from all the previous dense blocks. Following a $1\times1$ convolution, a softmax layer is used to provide the salient region maps.

Because the FCN-based salient region segmentation result is coarse and does not delineate the object borders, we use the fully connected CRFs to refine the segmentation prediction results \cite{krahenbuhl2011efficient}. Each pixel in the salient object map is further finely classified into labels of salient region or background by employing the energy function of CRF:
\begin{equation}
    E(S)=-\sum_{i}log P(s_i )+\sum_{i,j}\varphi_{p}(s_i,s_j),
\end{equation}

where $S$ represents a salient object map for all pixels, and $P(s_i)$ is the probability of pixel $x_i$ belonging to the label $s_i$, which indicates the saliency likelihood of pixel $x_i$. The pairwise cost $\varphi_p(s_i,s_j)$ for two labels $s_i$ and $s_j$ is defined as:

\begin{equation}\label{crfcost}
\begin{aligned}
\varphi _{p}\left( s_{i},s_{j}\right) =\omega _{1}\exp \left( -\dfrac {\left| p_{i}-p_{j}\right| ^{2}}{2\theta ^{2}_{\alpha}}-\dfrac {\left| I_{i}-I_{j}\right| ^{2}}{2\theta ^{2}_{\beta }}\right) +\\ w_{2}\exp \left( -\dfrac {\left| p_{i}-p_{j}\right| }{2\theta ^{2}_{\gamma}}\right) ^{2}
\end{aligned}
\end{equation}

where $\omega_{1}$ and $\omega_{2}$ indicate the relative weights corresponding to the two parts of \equref{crfcost}. The first part represents the appearance kernel, and the second part quantifies the smoothness kernel. The hyperparameters $\theta_{\alpha}$, $\theta_{\beta}$ and $\theta_{\gamma}$ are the standard deviation values that control the Gaussian kernels; $p_{i}$ and $p_{j}$ are the position vectors; and $I_{i}, I_{j}$ are the respective RGB vectors of the pixels $x_{i}$ and $x_{j}$. In this paper, we set the values of $\omega_{1}, \omega_{2}, \theta_{\alpha}, \theta_{\beta}$ and $\theta_{\gamma}$ to 30, 30, 61, 13 and 1, respectively. The model causes the fully connected CRF to enforce the structural consistency of the segmentation output and refines the areas to generate smoother contours \cite{chen2014semantic}. As shown in the example in \figref{DFCNnetwork}, compared to the prediction map, the handle of the handbag is both finer and more accurate after the CRF process. It is essential to detect the salient regions precisely to improve the next instance segmentation.

\subsection{Salient instance clustering}

When a known number of instances exist, it is natural to consider using a clustering algorithm to perform salient instance segmentation. Spectral clustering based on the graph model can produce more accurate and effective results than other, simpler clustering algorithms \cite{zelnik2005self}. The number of instances predicted by the DSN indicates a reasonable cluster number for the $k$ value used in spectral clustering. In this phase, we utilize the features extracted from the DFCN during the spectral clustering instead of the low-level features. In addition, the pixels in the input image are replaced by the superpixels generated by the SLIC algorithm and used during the clustering process \cite{achanta2012slic}. In general, k-means clustering is the last procedure in spectral clustering. Considering the drawbacks of using the local minimum to select the initial clustering centers, we also use fractile points to determine the initial clustering centers instead of simply making random selections.

First, we overlay the original image with the salient region map and use that as the input image to isolate the salient regions and mitigate background influence. Second, we found that the instances are segmented integrally and simply when the number of superpixels is set to between 200 and 300 \cite{long2015fully}; consequently, all the pixels in the input image are classified into superpixels with a size of 250. After the preprocessing step, we construct a single graph $G= (V, E)$, where $V$ consists of the nodes in the input image and $E$ is a set of undirected edges \cite{yang2013saliency}. The affinity matrix is defined by
\begin{equation}\label{affinitymatrix}
    \omega_{ij}=\dfrac {e^{-\dfrac {\left\| c_{i}-c_{j}\right\| }{\sigma ^{2}}}}{1+\lambda \cdot \left| \left| d_{i}-d_{j}\right| \right| }\,\,\,\,\,\,\,\,i,j\in V,
\end{equation}
where $\lambda$ is a parameter to control the spatial distance, which is set to 3 in our implementation. $\left\| d_{i}-d_{j}\right\|$ is the Euclidean distance between pixel $i$ and $j$. The parameter $\sigma^2$ controls how rapidly the affinity $\omega_{ij}$ declines with the distance between $c_{i}$ and $c_{j}$, and we set $\sigma^2$ to 10 \cite{huang2017densely}. $c_{i}$ and $c_{j}$ denote the means of the superpixels corresponding to two nodes in the feature map produced by DFCN. We extract 6 candidate deep features in DFCN and resize those feature maps so that their size corresponds with the input image; the locations of 6 feature maps are shown in \figref{DFCNnetwork}. We discussed the effectiveness of different feature maps working on the spectral clustering in Section \uppercase\expandafter{\romannumeral4}$\textbf{-}$B. Given image $X$, where all the superpixels are represented by $(v_{1},v_{2},…,v_{n})$, the degree matrix is $D=diag\{d_{11},…,d_{nn}\}$, $d_{ii}=\sum_{j}\omega_{ij}$. In conjunction with $D$, we find the smallest $k$ eigenvalues using $D^{-1/2} LD^{-1/2}$ and then generate the corresponding eigenvectors $U$.

Finally, k-means clustering is used to cluster the eigenvectors $U$ and obtain the salient instance segmentation results. The traditional k-means clustering algorithm randomly selects $k$ pixel points from an image to be the initial cluster centers. However, randomly selecting initial cluster centers can cause the clustering process to become stuck in local minima, which prevents the salient instances from being completely segmented. Therefore, we use the fractile concept to find the initial cluster centers and avoid the negative influence of random selection on the final results. Following statistical methods, we arrange all the values from small to large and divide them into four equal parts. The values at the three split points are the quartiles. Thus, we divide all the pixels of the eigenvectors $U$ into $k$ equal parts from small to large and adopt the center point of each part as the initial cluster center. The method provides some guidance when selecting the initial cluster centers. First, we arrange all the pixels of an image in ascending order based on their values into the characteristic matrix $U$. Second, we let $Q$ be a fractile of this vector, where $Q=Q_{1},Q_{2},…,Q_{i}$. The value of $Q_{i}$ is calculated as follows:

\begin{equation}
    Q_{i}=\left[ \dfrac {50}{k}+\left( i-1\right) \dfrac {100}{k}\right] \cdot U,i=1,2,\ldots ,k,
\end{equation}

where $k$ is the number of clusters, and $U$ is the vector of the characteristic matrix arranged by ascending values. When the value of $k$ is 4, we will obtain four fractiles: $Q_{1}$,$Q_{2}$,$Q_{3}$,and $Q_{4}$. These $Q$ values are what we adopt for the cluster centers. It is both convenient to assign the initial pixels through this initial guidance method and it also enhances the stability of salient instance segmentation. The major procedures by adaptive spectral clustering method are summarized in Algorithm 1.

\begin{algorithm}
    \renewcommand{\algorithmicrequire}{\textbf{Input:}}
    \renewcommand{\algorithmicensure}{\textbf{Output:}}
    \caption{Salient Instance Clustering}
    \label{alg:1}
    \begin{algorithmic}[1]
        \REQUIRE An image, the corresponding salient object map and a subitizing $k$
        \STATE Refine the image to filter out the background depending on the saliency map.
        \STATE Segment the image produced in step 1 into superpixels and build a graph $G$.
        \STATE The feature map extracted by the DFCN is embed to compute the degree matrix $D$ and affinity matrix $W$ using \equref{affinitymatrix}.
        \STATE Calculate the normalized Laplace matrix and add the number of instances k to obtain the corresponding eigenvectors $U$.
        \STATE Use the improved k-means clustering to classify the eigenvectors $U$ and resize the label map by the pixel list in the SLIC algorithm to obtain the salient instance results.
        \ENSURE The salient instance segmentation map, in which each instance is represented by one label.
    \end{algorithmic}
\end{algorithm}

\subsection{Implementation Details}

In this section, we present more details regarding the training phase of MDNN. We employed a weighted cross-entropy function as the loss function to train the network, which is actuated as follows:

\begin{equation}
     L\left( y,\widehat {y}\right) =-\dfrac {1}{N}\sum ^{N}_{i=1}\sum ^{c}_{c=1}y_{i}^{c}\log \widehat{y}_{i}^{c},，
\end{equation}

where $\hat{y}_{i}^{c}$ denotes the probability of pixel $i$ belonging to class $c$ ($c=4$ in the DSN and $c=2$ in the DFCN), and $y_{i}^{c}$ indicates the ground truth label for pixel $i$. Because we use separate training sets for the two subtask networks, the parameters of these two networks cannot be shared. Both networks are trained using stochastic gradient descent (SGD) \cite{lecun1989backpropagation}. During the training phase, the weight decay is empirically set to $5\times10^{-4}$, and the momentum is 0.95 without dampening. For the DSN, the initial learning rate is set to 0.001, which is then divided by a factor of 0.1 every 7 epochs. The weight decay is applied to the weights in the convolution and fully connected layers. In addition, we use a mini-batch size of 8 for 100 epochs. In DFCN, the learning rate is set to $10^{-7}$, and the weight decay is applied only to the convolution weights. Different from the DSN, the mini-batch size is set to 6 for 100 epochs. The entire procedure is repeated iteratively for training.

\section{EXPERIMENTS}

In this section, we thoroughly explain the details of the experimental process and evaluate the results of various procedures on different performance metrics, including subitizing, salient region detection and salient instance segmentation tasks. To demonstrate the validity of our approach, we execute our MDNN method on several public datasets and compare it equitably with the state-of-the-art methods equitably. In addition, we analyze the different strategies used in our approach and adopt the optimal solutions in the subsequent experiments. Finally, we discuss a set of experimental results using qualitative and quantitative criteria.

\subsection{Experimental settings}

As described in Section \uppercase\expandafter{\romannumeral3}$\textbf{-}$C, our proposed MDNN is implemented in the PyTorch framework on 2 NVIDIA GeForce GTX 1080Ti GPUs with 22 GB of memory. Because none of the existing image datasets contain both subitizing and salient object maps, the subnetworks cannot all be trained together; thus, their parameters cannot be optimized simultaneously. In the experiment, we use 500 images for training, 200 for validation and 300 for testing from the index of dataset1k. Before using the dataset1k, we labeled the number of instances for the subitizing task to reflect the ground-truth. Because the sole available salient instance segmentation dataset1k contains only 500 images for training, we extended the training set by adding the datasets SOS \cite{zhang2017salient} and MSRA-B \cite{yan2013hierarchical} for the subitizing and salient region detection subnetworks, respectively. For the DSN, the SOS dataset, which includes 13,707 images proposed by Zhang \etal \cite{zhang2017salient}, was added for training. For fair comparison, we generated synthetic images to train the CNN model before fine-tuning on the real data, which is same as \cite{zhang2017salient}. Besides, we resized all the images and ground-truth maps to $256\times256$ and randomly cropped the region into a $224\times224$ square image. During validation, the images were resized to $224\times224$ regardless of their original aspect ratios. For the DFCN, we combined the MSRA-B (5000 images) into the training sets \cite{yan2013hierarchical}. Different from the previous network, we resized the images to $340\times340$ and randomly cropped the regions into $300\times300$ square images. The MDNN is fine-tuned by flipping the training sets horizontally at a probability of 0.5, and we did not perform any random augmentations during validation.

Training the two subtask networks takes approximately 7 hours for the DSN and approximately 10 hours for the DFCN. During the test phase, it requires approximately 0.8 seconds per image to produce a salient region map and determine the number of instances and another 0.5 seconds to segment the instances using the proposed spectral clustering on a $340\times340$ test image. The overall time cost is 1.3 seconds per image, which is considerably less than MSRNet \cite{li2017instance}, which requires more than 20 seconds per image.

\begin{figure}[!t]
    \centering
    \includegraphics[width=0.6\linewidth]{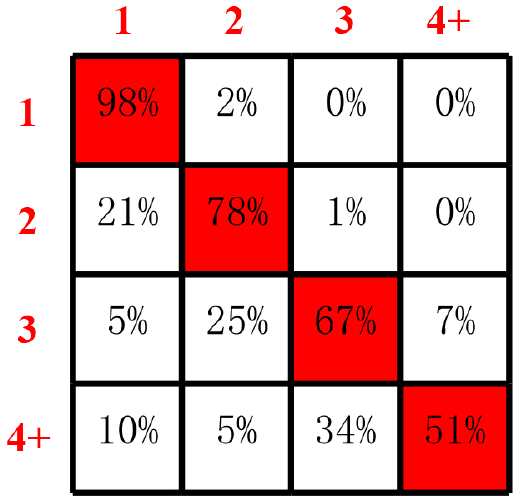}
    \caption{Confusion matrix of the proposed DSN. Each cell includes the accuracy percentage (recall). Each row corresponds to the ground-truth number, while each column shows the predicted results. The values on the diagonal line represent the correct result percentages.}
    \label{Confusion matrixDSN}
    \vspace{-0.1in}
\end{figure}

\begin{table}[!t]
    \renewcommand{\arraystretch}{1.3}
    \caption{Comparison of the DSN and different versions of DenseNet model. The second row represents the average precision $(\%)$ of the predicted number of instances.}
    \label{tab:Densenet}
    \centering
    \resizebox{0.48\textwidth}{!}{
\begin{tabular}{c|c|c|c|c}
\hline
DSN  & DenseNet-121 & DenseNet-161 & DenseNet-169 & DenseNet-201 \\ \hline
79.4 & 68.7         & 72.5         & 70.3         & 77.5         \\ \hline
\end{tabular}}
\end{table}

\begin{table}[!t]
\renewcommand{\arraystretch}{1.3}
\caption{The AP score performance of DSN with different connection types of SE block. Inter-SE means SE block is appended to head and tail of each dense block and Intra-SE means SE block is connected inside the dense block. The highest scores in each row are labeled in bold.}
    \label{SE-connection}
    \centering
    \setlength{\tabcolsep}{3.3mm}{
\begin{tabular}{c|c|c|c|c|c|c}
\hline
Method   & 0             & 1             & 2             & 3             & 4+            & mean          \\ \hline
intra-SE & 98.1          & 96.7          & 80.2          & 72.7          & 77.5          & 85            \\ \hline
inter-SE & \textbf{98.3} & \textbf{97.1} & \textbf{82.4} & \textbf{75.3} & \textbf{76.4} & \textbf{85.9} \\ \hline
\end{tabular}}
\end{table}

\begin{table}[!t]
\renewcommand{\arraystretch}{1.3}
    \caption{Quantitative comparisons with the results obtained by varying the feature maps in the spectral clustering algorithm. The bold values indicate the best performance.}
   \label{featurecompare}
   \centering
   \resizebox{0.48\textwidth}{!}{
\begin{tabular}{c|c|c|c|c|c|c}
\hline
Metrics                           & Feature1 & Feature2 & Feature3 & Feature4 & Feature5 & Feature6       \\ \hline
$AP^{r}@0.5(\%)$ & 50.89    & 53.18    & 57.46    & 62.74    & 69.24    & \textbf{73.46} \\ \hline
$AP^{r}@0.7(\%)$ & 38.2     & 40.66    & 43.83    & 48.96    & 56.18    & \textbf{60.14} \\ \hline
\end{tabular}}
\end{table}

\begin{table*}[!t]
\renewcommand{\arraystretch}{1.3}
    \caption{Comparison of maximum F-measure (larger is better), MAE scores (smaller is better), S-measure (larger is better) and E-measure (larger is better). The best three scores in each row are shown in red, blue, and green, respectively.}
    \label{ComparisonSOD}
    \centering
    \resizebox{\textwidth}{!}{
\begin{tabular}{c|c|c|c|c|c|c|c|c|c|c|c|c|c|c|c}
\hline
Dataset                     & Metric   & GC    & GMR    & DRFI                         & FT    & BMS   & MDF   & MTDNN                        & DCL+                         & MSRNet                       & DSS                          & PAGR                         & R3Net                        & OURS                         & OURS-CRF                     \\ \hline
                            & maxF     & 0.537 & 0.61   & 0.665                        & 0.313 & 0.516 & 0.68  & 0.745                        & 0.757                        & 0.785                        & 0.764                        & 0.711                        & {\color[HTML]{32CB00} 0.786} & {\color[HTML]{34CDF9} 0.788} & {\color[HTML]{FE0000} 0.795} \\ \cline{2-16} 
                            & MAE      & 0.168 & 0.141  & 0.098                        & 0.249 & 0.174 & 0.115 & 0.076                        & 0.08                         & 0.069                        & 0.07                         & 0.072                        & {\color[HTML]{32CB00} 0.063} & {\color[HTML]{34CDF9} 0.051} & {\color[HTML]{FE0000} 0.05}  \\ \cline{2-16} 
                            & Smeasure & 0.598 & 0.624  & 0.655                        & 0.503 & 0.624 & 0.721 & 0.75                         & 0.771                        & 0.808                        & 0.784                        & 0.783                        & {\color[HTML]{32CB00} 0.817} & {\color[HTML]{FE0000} 0.827} & {\color[HTML]{34CDF9} 0.824} \\ \cline{2-16} 
\multirow{-4}{*}{DUT-OMRON} & Emeasure & 0.681 & 0.725  & 0.748                        & 0.616 & 0.725 & 0.783 & 0.835                        & 0.815                        & {\color[HTML]{32CB00} 0.86}  & 0.838                        & 0.724                        & 0.857                        & {\color[HTML]{FE0000} 0.888} & {\color[HTML]{34CDF9} 0.883} \\ \hline
                            & maxF     & 0.624 & 0.7012 & 0.739                        & 0.38  & 0.63  & 0.807 & 0.81                         & 0.892                        & {\color[HTML]{34CDF9} 0.913} & {\color[HTML]{32CB00} 0.908} & 0.891                        & {\color[HTML]{FE0000} 0.935} & 0.873                        & 0.887                        \\ \cline{2-16} 
                            & MAE      & 0.238 & 0.204  & 0.184                        & 0.327 & 0.213 & 0.138 & 0.122                        & 0.068                        & {\color[HTML]{32CB00} 0.054} & 0.064                        & 0.064                        & {\color[HTML]{FE0000} 0.04}  & {\color[HTML]{32CB00} 0.054} & {\color[HTML]{34CDF9} 0.052} \\ \cline{2-16} 
                            & Smeasure & 0.559 & 0.606  & 0.732                        & 0.454 & 0.581 & 0.776 & 0.821                        & 0.868                        & {\color[HTML]{34CDF9} 0.895} & 0.864                        & {\color[HTML]{32CB00} 0.889} & {\color[HTML]{FE0000} 0.91}  & 0.866                        & 0.869                        \\ \cline{2-16} 
\multirow{-4}{*}{ECSSD}     & Emeasure & 0.63  & 0.664  & 0.752                        & 0.558 & 0.654 & 0.776 & 0.913                        & 0.823                        & 0.937                        & 0.846                        & {\color[HTML]{34CDF9} 0.928} & {\color[HTML]{FE0000} 0.929} & {\color[HTML]{32CB00} 0.926} & {\color[HTML]{34CDF9} 0.928} \\ \hline
                            & maxF     & 0.719 & 0.777  & 0.845                        & 0.579 & 0.742 & 0.885 & 0.845                        & 0.916                        & {\color[HTML]{FE0000} 0.93}  & {\color[HTML]{34CDF9} 0.92}  & 0.913                        & 0.915                        & 0.916                        & {\color[HTML]{32CB00} 0.918} \\ \cline{2-16} 
                            & MAE      & 0.159 & 0.128  & 0.112                        & 0.241 & 0.152 & 0.066 & 0.094                        & 0.047                        & 0.042                        & 0.043                        & {\color[HTML]{32CB00} 0.04}  & {\color[HTML]{34CDF9} 0.039} & {\color[HTML]{34CDF9} 0.039} & {\color[HTML]{FE0000} 0.037} \\ \cline{2-16} 
                            & Smeasure & 0.701 & 0.753  & 0.804                        & 0.582 & 0.728 & 0.871 & 0.816                        & 0.891                        & 0.894                        & 0.887                        & 0.884                        & {\color[HTML]{32CB00} 0.895} & {\color[HTML]{FE0000} 0.899} & {\color[HTML]{34CDF9} 0.898} \\ \cline{2-16} 
\multirow{-4}{*}{MSRA-B}    & Emeasure & 0.725 & 0.796  & 0.891                        & 0.678 & 0.746 & 0.902 & 0.849                        & 0.934                        & {\color[HTML]{34CDF9} 0.941} & {\color[HTML]{32CB00} 0.938} & 0.935                        & 0.936                        & {\color[HTML]{FE0000} 0.943} & {\color[HTML]{FE0000} 0.943} \\ \hline
                            & maxF     & 0.734 & 0.767  & {\color[HTML]{34CDF9} 0.839} & 0.633 & 0.642 & 0.823 & {\color[HTML]{FE0000} 0.841} & 0.831                        & {\color[HTML]{32CB00} 0.837} & 0.83                         & 0.780.                       & 0.835                        & 0.832                        & 0.833                        \\ \cline{2-16} 
                            & MAE      & 0.18  & 0.1567 & 0.123                        & 0.203 & 0.172 & 0.115 & 0.124                        & {\color[HTML]{FE0000} 0.078} & {\color[HTML]{34CDF9} 0.08}  & 0.098                        & 0.103                        & 0..091                       & 0.082                        & {\color[HTML]{32CB00} 0.081} \\ \cline{2-16} 
                            & Smeasure & 0.706 & 0.74   & 0.774                        & 0.693 & 0.7   & 0.791 & 0.776                        & 0.757                        & {\color[HTML]{32CB00} 0.812} & 0.771                        & 0.757                        & 0.808                        & {\color[HTML]{34CDF9} 0.814} & {\color[HTML]{FE0000} 0.815} \\ \cline{2-16} 
\multirow{-4}{*}{SED2}      & Emeasure & 0.737 & 0.781  & 0.822                        & 0.755 & 0.736 & 0.84  & 0.825                        & 0.851                        & 0.873                        & 0.868                        & 0.853                        & {\color[HTML]{34CDF9} 0.876} & {\color[HTML]{32CB00} 0.875} & {\color[HTML]{FE0000} 0.878} \\ \hline
                            & maxF     & 0.555 & 0.642  & 0.677                        & 0.441 & 0.552 & 0.785 & 0.753                        & 0.832                        & 0.799                        & {\color[HTML]{32CB00} 0.836} & 0.761                        & 0.816                        & {\color[HTML]{34CDF9} 0.837} & {\color[HTML]{FE0000} 0.841} \\ \cline{2-16} 
                            & MAE      & 0.252 & 0.23   & 0.216                        & 0.284 & 0.262 & 0.155 & 0.179                        & 0.126                        & {\color[HTML]{32CB00} 0.114} & 0.126                        & 0.147                        & 0.124                        & {\color[HTML]{34CDF9} 0.112} & {\color[HTML]{FE0000} 0.111} \\ \cline{2-16} 
                            & Smeasure & 0.582 & 0.608  & 0.619                        & 0.497 & 0.574 & 0.674 & 0.714                        & 0.747                        & {\color[HTML]{FE0000} 0.782} & 0.746                        & 0.72                         & {\color[HTML]{34CDF9} 0.761} & 0.748                        & {\color[HTML]{32CB00} 0.752} \\ \cline{2-16} 
\multirow{-4}{*}{SOD}       & Emeasure & 0.696 & 0.717  & 0.728                        & 0.61  & 0.681 & 0.752 & 0.761                        & 0.797                        & 0.803                        & 0.804                        & {\color[HTML]{32CB00} 0.811} & {\color[HTML]{FE0000} 0.835} & 0.807                        & {\color[HTML]{34CDF9} 0.813} \\ \hline
\end{tabular}}
\end{table*}

\subsection{Results and comparisons}

\textbf{Ablation Study:} To investigate the effectiveness of different structures in our method, we conduct the ablation study. First, to verify the most appropriate basic network structures, we trained four candidate models based on DenseNet \cite{huang2017densely}, named DenseNet-121, DenseNet-161, DenseNet-169 and DenseNet-201. These models have different numbers of layers in the dense blocks \cite{huang2017densely}. All the compared models are trained with the same settings as our DSN. Their average precision (AP) results are listed in \tabref{tab:Densenet}, which shows that our DSN achieves the best performance in terms of the AP on the test images. DenseNet-121 performs the worst because it has fewer layer in its dense blocks. The other two models exhibit similar performances behind our network. DenseNet-201 achieved $77.5\%$ of the AP scores which is better than other version of DenseNet, consequently we choose DenseNet-201 as the backbone of DSN.

The good performance of DSN is also benefited by inserting the SE block. In order to further improve the performance, we experiment two ways to embed the SE blocks in DSN: insert them within the dense block or append them to head and tail of each dense block. The results of different connection modes using SOS dataset are displayed in \tabref{SE-connection}. We can see that the connection mode of inter-SE is generally better than the intra-SE mode no matter how many instances in a scene. Finally, we choose the inter-SE mode that make the SE block is appended to head and tail of each dense block in DSN.

In addition, to evaluate the performance of different feature maps in the spectral clustering operation, we also test six feature maps extracted from DFCN to generate the final salient instance segmentation. The marked number of feature maps corresponds to \figref{DFCNnetwork}, feature map 1, 2 and 3 belongs to the front of DFCN and feature map 4, 5 and 6 comes from the latter layers of DFCN. As we know that the first few layers of features extracted by the DFCN includes relatively more low-level features, while latter maps contain more high-level features. In this experiment, we added these feature maps in the spectral clustering to obtain the performance metrics $AP^{r}$@0.5 and $AP^{r}$@0.7 (\equref{APR0.5}) reported in \tabref{featurecompare}. \tabref{featurecompare} shows that the feature map 6 achieves the beat results. It also demonstrates that the latter feature maps have higher $AP^{r}$ scores than the former features. The reason is the input feature to the dense block in DFCN consists of the summed information from all the previous dense blocks, so the latter map contains more abundant feature including low-level and high-level features.

\textbf{Evaluation of subitizing:} To evaluate the number of instances, we first input 300 test images from the index of dataset1k to the DSN. To demonstrate the accuracy of the predicted subitizing, \figref{Confusion matrixDSN} shows a confusion matrix for the distribution of the results by the DSN. Different from SOS [20], the DSN excludes the number 0 because dataset1K have no image without a salient object. The matrix presents the percentage of results compared to the ground truth. The recall value is stable and is as expected when the number of instances is below 4. The accuracy rate for category 1 was the highest (98\%). The recall values for categories 2 and 3 are stable and exhibit a gradual downward trend. When the images contain 4 or more objects, the accuracy decreases to (51\%). It is interesting to note that most incorrect results are located near correct results. This finding demonstrates that we can continue to segment the corresponding number of instances while losing few instances. It is worth noting that the prediction ratio for category 1 in the 4+ images is (10\%), which can be interpreted as one object of the image being substantially larger than the other 3 objects, causing the algorithm to fail to count them correctly.

\begin{table}[!t]
    \renewcommand{\arraystretch}{1.3}
    \caption{Average Precision comparison (\%) of the subitizing task. Category 0 was added for this comparison, and the separated and mean of the AP scores are reported. The highest scores in each row are labeled in bold.}
    \label{APsubitizing}
    \centering
\begin{tabular}{c|c|c|c|c|c|c}
\hline
Method       & 0             & 1             & 2             & 3             & 4+            & mean          \\ \hline
CNN-Syn-FT \cite{zhang2017salient} & 93.5          & 93.8          & 77.4          & 64.3          & 73            & 80.4          \\ \hline
DSN          & \textbf{98.3} & \textbf{97.1} & \textbf{82.4} & \textbf{75.3} & \textbf{76.4} & \textbf{85.9} \\ \hline
\end{tabular}
\end{table}

To further demonstrate the efficiency of our algorithm, we tested the DSN on the SOS dataset (with 13,707 images) compared with the CNN-Syn-FT model provided by Zhang \etal \cite{zhang2017salient}.To ensure a fair comparison, we used the two-stage fine-tuning scheme with the real and synthetic image data to train DSN. For pre-training using the synthetic images, we generated 34,000 for each number in 1$-$4 as same as \cite{zhang2017salient}. In addition, DSN training and testing parameters were conducted in line with those of the CNN-Syn-FT. \tabref{APsubitizing} shows the average precision scores. We can see that DSN is comprehensive superior to CNN-Syn-FT and the mean AP scores of our network is 5 percentage points better than CNN-Syn-FT. Especially, its accuracy scores on category 3 is higher than CNN-Syn-FT about 11 percent.

\textbf{Evaluation of salient region detection:} The salient region detection task has a direct impact on the salient instance segmentation results. To validate the performance of DFCN, we conducted experiments on four publicly available benchmark datasets annotated with pixelwise ground-truth labeling: DUT-OMRON \cite{yang2013saliency}, ECSSD \cite{yan2013hierarchical}, MSRA-B \cite{liu2011learning}, SOD \cite{movahedi2010design} and SED2 \cite{movahedi2010design}. For these experiments, we adopted four of the most representative evaluation metrics to evaluate the salient region maps. The first metric is the F-measure, which evaluates the quality of salient region maps after binarizing them using a specific threshold. The F-measure is computed as follows:
\begin{equation}
    F_{\beta} =\frac{(1+\beta^2 )Precision×Recall}{\beta^2 Precision+Recall},
\end{equation}
and $\beta^{2}$ was set to 0.3 \cite{lin2018saliency}. In this experiment, we used the maximum F-measure (maxF) instead of the average F-measure (aveF) because of the adaptive threshold. The second metric is mean absolute error (MAE), which reflects the mean pixelwise difference between the saliency map and the ground truth \cite{borji2015salient} and is calculated as
\begin{equation}
    MAE=\frac{1}{W\times H}\sum_{x=1}^{W}\sum_{y=1}^{H}\left \| S(x,y)-G(x,y) \right \|,
\end{equation}
where $S$ and $G$ denote the saliency map and the ground truth, respectively. $W$ and $H$ are the image size parameters. This metric of assessing the salient regions compared with the ground truth is intended to reflect the salient instance detection quality. A smaller MAE value denotes better quality prediction results. The two other metrics are structure measure (S-measure) \cite{fan2017structure} and Enhanced measure (E-measure) \cite{Fan2018Enhanced}, which have added to serve as the supplement of the F-measure and MAE. The higher that S-measure and E-measure are, the better results are the methods.

For comparison with other salient region detection methods, we performed a horizontal evaluation with 12 classic or state-of-the-art methods, including GC \cite{cheng2013efficient}, GMR \cite{lecun1989backpropagation}, DRFI \cite{jiang2013salient}, FT \cite{yan2013hierarchical}, BMS \cite{zhang2013saliency}, MDF \cite{li2015visual}, MTDNN \cite{li2016deepsaliency}, DCL \cite{li2016deep}, MSRNet \cite{li2017instance}, DSS \cite{hou2017deeply}, PAGR \cite{zhang2018progressive} and \cite{deng2018r3net}. Among these methods, the first five are implemented based on traditional machine learning, while the rest of  approaches operate using deep learning models. We conducted the experiments by executing the publicly available source code provided by the original authors. \tabref{ComparisonSOD} shows the results from our DFCN compared with those of the other methods on the metrics w.r.t maxF, MAE, S-measure and E-measure on five benchmark datasets. Clearly, DFCN performs stably and favorably when compared to the state-of-the-art methods in most cases. More specifically, the proposed network performs best with regard to the MAE value, which is highly important for the salient instance detection task. On a macro level, the deep-learning approaches achieve better performances than do the traditional machine learning methods. In addition, the last column in \tabref{ComparisonSOD} reports the results after CRF \cite{krahenbuhl2011efficient}, which comprehensively improves the earlier results measurements whether considering any one of the metrics; however, the magnitudes of the increases are not large.

\begin{table}
\renewcommand{\arraystretch}{1.3}
    \caption{Quantitative comparisons with existing methods by the $AP^{r}$ metric at IoU threshold of 0.5 to 0.9. The bold values indicate the best performances.}
   \label{ComparisonSID}
   \centering
   \resizebox{0.48\textwidth}{!}{
\begin{tabular}{c|c|c|c|c|c}
\hline
Method & AP$^{r}$0.5(\%)      & AP$^{r}$0.6(\%)     & AP$^{r}$0.7(\%)      & AP$^{r}$0.8(\%)      & AP$^{r}$0.9(\%)      \\ \hline
MSRNet & 65.32          & -             & 52.18          & -              & -              \\ \hline
S4Net  & \textbf{74.16} & 66.21         & 55.34          & 36.95          & 12.74          \\ \hline
Ours   & 73.46          & \textbf{67.3} & \textbf{60.14} & \textbf{46.25} & \textbf{23.51} \\ \hline
\end{tabular}}
\end{table}

\begin{table}
\renewcommand{\arraystretch}{1.3}
    \caption{Comparing the $AP^{r}$ metrics for different numbers of instances per image found by the existing methods on dataset1K. The bold values indicate the best performances.}
    \label{tab:SIDin1kdataset}
    \centering
    \resizebox{0.48\textwidth}{!}{
\begin{tabular}{c|c|c|c|c|c}
\hline
Method & Metrics   & 1     & 2     & 3     & 4+    \\ \hline
S4Net  & AP$^{r}$0.5(\%) & 81.98 & 74.70  & 63.71 & \textbf{49.25} \\ \hline
Ours   & AP$^{r}$0.5(\%) & \textbf{83.44} & \textbf{75.20} & \textbf{64.23} & 43.22 \\ \hline\hline
S4Net  & AP$^{r}$0.7(\%) & 65.64 & 54.24  & 47.57 & 31.54 \\ \hline
Ours   & AP$^{r}$0.7(\%) & \textbf{70.09} & \textbf{58.08} & \textbf{58.53} & \textbf{32.50}  \\ \hline
\end{tabular}}
\end{table}

\textbf{Evaluation of salient instance segmentation:} Because salient instance segmentation is a new and challenging task, few performance measures exist to perform a quantitative verification. Therefore, to evaluate the effectiveness of our approach for salient instance segmentation, we refer to the metrics for semantic segmentation provided by \cite{hariharan2014simultaneous}. Different from the mean average precision (MAP) for semantic segmentation, salient object instances are independent of category; consequently, we adopt the $AP^{r}$ metric instead of the $MAP^{r}$ metric at intersection-over-union (IoU) scores of 0.5 to 0.9. Taking the threshold of IoU at 0.5 as an example, we first need to calculate the per image precision of each instance from the dataset. Fixing the precision value at $IoU>0.5$, the sum of these precision values divided by the number of all ground-truth instances obtains the $AP^{r}$ at 0.5 IoU scores. This operation is followed by the following calculation:
\begin{equation}\label{APR0.5}
    AP^{r}@0.5=\frac{1}{N}\sum_{i=1}^{K} precision(i)\,\,\,\,\,\,\,\,IoU(i)\geq 0.5,
\end{equation}
where $K$ is the number of instances in which the value of IoU is greater than 0.5, and $N$ denotes the total number of ground-truth instances in the dataset.

Due to datasets limitations, we adopted only the dataset including 1,000 images provided by \cite{li2017instance}. We used the published results on the dataset1K test set directly; all the other settings were the same as those used in \cite{li2017instance} to ensure a fair comparison. We report the results of the $AP^r$ of the metrics with IoU scores of 0.5 to 0.9 predicted by the state-of-the-art methods in \tabref{ComparisonSID}. The proposed method performs better than MSRNet \cite{li2017instance} and S4Net \cite{fan2019s4net} when segmenting the salient instances, especially when the $AP^{r}$ metric is greater than the IoU score of 0.6. due to the related code of \cite{li2017instance} is not available, we cannot get its results of $AP^{r}$ metric.

The detailed $AP^{r}$ scores for different numbers of instances per image by S4Net \cite{fan2019s4net} and our method are listed in \tabref{tab:SIDin1kdataset}, it reveals that the performances showed a downward trend as the number of instances in the images increased. Comparing with S4Net, our results outperform against them generally, while the $AP^{r}$@0.5 score  for the number of 4+ instances is smaller than \cite{fan2019s4net}. We can clearly see that using the Non-maximum Suppression (NMS) method to estimate the number of proposals is not stable enough compared with DSN. The \tabref{tab:SIDin1kdataset} also illustrates that the accuracy rate of the subitizing prediction has a positive influence on the salient instance segmentation. Overall, the performance of MDNN is as expected and it outperforms the state-of-the-art methods.

\begin{figure}[!t]
    \centering
    \includegraphics[width=0.85\linewidth]{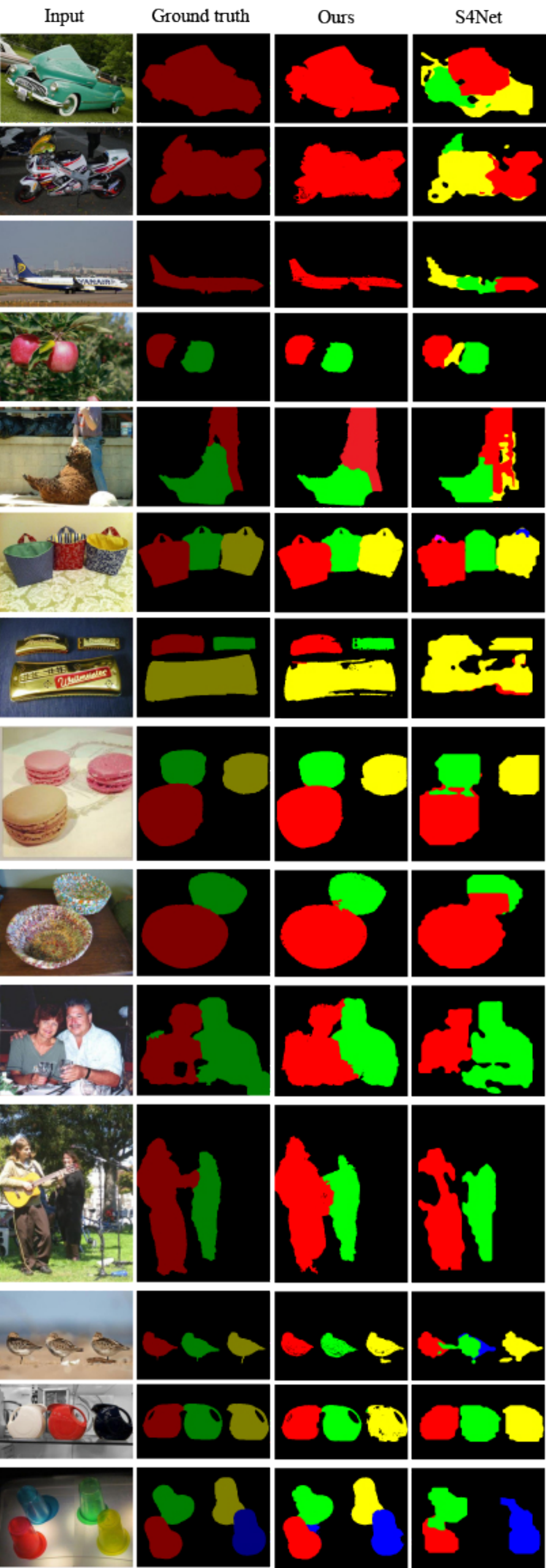}
    \caption{Examples of salient instance segmentation results by the proposed method and S4Net \cite{fan2019s4net}. In each image, the different colors indicate different instances.}
    \label{Examplesofourmethod}
    \vspace{-0.1in}
\end{figure}

We also qualitatively evaluated our method on the salient instance segmentation dataset. \figref{Examplesofourmethod} visualizes the predicted salient instance segmentation results using our approach. Due to the lack of published code for the only related work (MSRNet \cite{li2017instance}), we were not able to visually compare the results with those of any existing method. Instead, we chose representative segmentation results for analysis in \figref{Examplesofourmethod}. To be consistent with the ground truth, the segmented instances are labeled with different colors corresponding to the ground truth. The proposed approach achieves satisfactory performance regardless of how complicated the original images are. The first three rows show instances that have similar internal features and are close together in one image. The fourth row shows two overlapping instances with complex color and texture features; however, the proposed method still segments them effectively. The middle rows show some images with messy backgrounds, and the instances are confused by the background. For example, in the fifth row, the woman is wearing a dark-colored coat and partially blends in with the tall tree behind her. In addition, the color of the sheep in the three last row image blends with the rocks and trees in the background. Other segmentation results reveal other challenging cases, including occlusions, diverse phenotypes and different views. The salient instances are segmented distinctly and consistently with the ground truth annotations.

\textbf{Visual Comparison:} We also qualitatively evaluated our method and S4Net \cite{fan2019s4net} on the salient instance segmentation dataset. \figref{Examplesofourmethod} visualizes the predicted salient instance segmentation results and we chose representative segmentation results for analysis. To be consistent with the ground truth, the segmented instances are labeled with different colors corresponding to the ground truth. The proposed approach achieves satisfactory performance regardless of how complicated the original images are. Compared to S4Net \cite{fan2019s4net}, our method gets the better segmentation results which are closer to the ground truth, especially when the number of instances is below 3. To be specific, the first three rows show the performance of different methods when the number of instances is 1. We can see it very intuitively that the proposed model can predict the number of instances accurately while S4Net generated unsatisfactory results because of estimating the number of instances incorrectly \cite{fan2019s4net}. The followed four rows show instances that have similar internal features and are close together in one image. The eighth and ninth rows show two overlapping instances with complex color and texture features; however, the proposed method still segments them effectively. The followed rows show some images with messy backgrounds, and the instances are confused by the background. For example, in the fourth from the bottom row, the woman is wearing a dark-colored coat and partially blends in with the tall tree behind her. In addition, the color of the birds in the three last row image blends with the rocks and trees in the background. Other segmentation results reveal other challenging cases, including occlusions, diverse phenotypes and different views. The salient instances of our method are segmented distinctly and consistently with the ground truth annotations. In contrast, the effectiveness of S4Net \cite{fan2019s4net} is slightly inferior to the proposed method. it reports that our novel method is highly suitable for solving the salient instance segmentation task. However, the lack of datasets id an urgent problem that limits the generalizability of our approach.

\begin{figure}[!t]
    \centering
    \includegraphics[width=\linewidth]{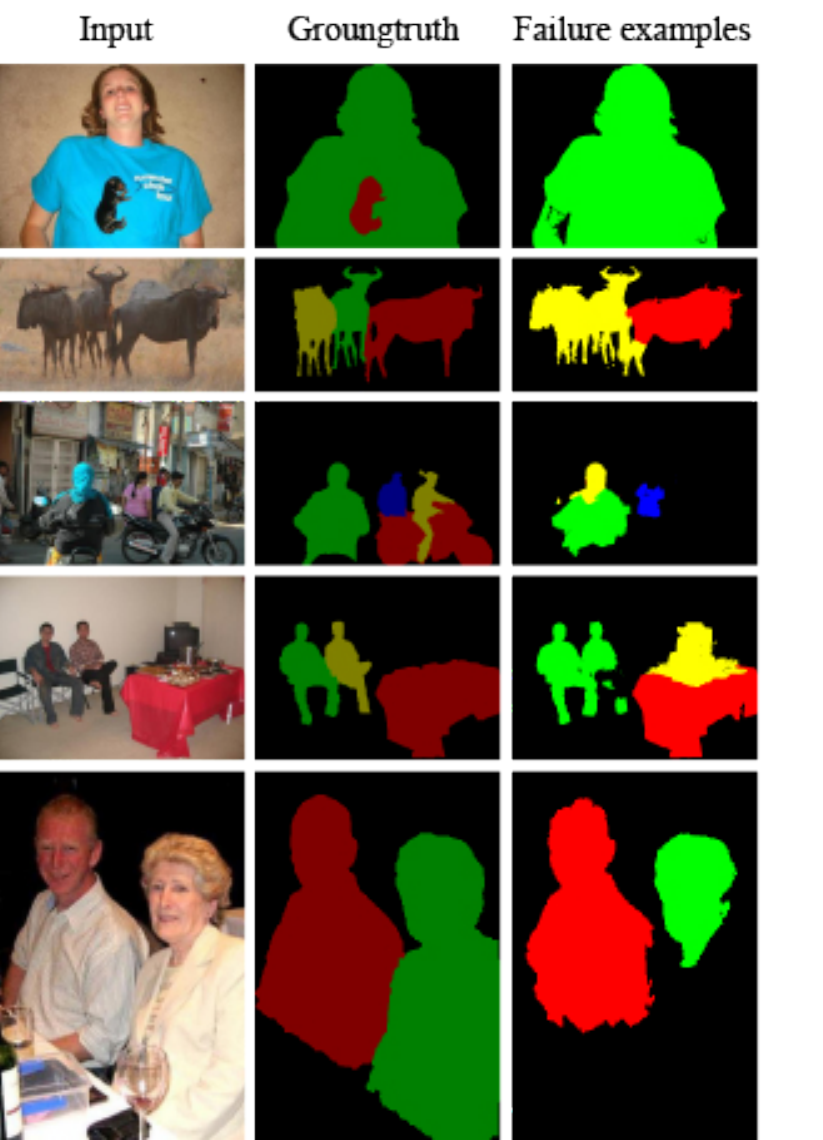}
    \caption{Illustration of failure examples generated by our method.}
    \label{failureexamples}
    \vspace{-0.1in}
\end{figure}

Some segmentation failure examples are shown in \figref{failureexamples}. These examples illustrate that some results were segmented unsatisfactorily because of an incorrect number of instances. Inaccurate salient regions can also lead to errors for the final instances, such as the results in the third and fifth rows. In addition, the image in the third row has diverse instance types, and the entire scene is chaotic, which can cause inexact results. When the instances are too small, the adaptive deep feature-based spectral clustering neglects them, which leads to segmentation failure. We plan to address these failure examples to handle tough instances in future work.

\section{CONCLUSION AND FUTURE WORKS}
Salient instance segmentation is a new and challenging task that is an extension of salient region detection. Thus far, limited research has been conducted for salient instance segmentation since the first work by Li \etal \cite{li2017instance}. In this paper, we introduced a proposal-free method that performs salient instance segmentation. Rather than using region proposal methods, the proposal-free network, named MDNN, incorporates the DSN and the DFCN. The MDNN can directly predict the salient regions and the number of instances in a category-independent manner. Adaptive spectral clustering is applied to the feature maps extracted from the DFCN to generate the final salient instances. The experimental results show that the proposed method achieves significant improvements for salient instance segmentation. More concretely, the subtask networks that perform saliency region detection and the subitizing perform well against most previous state-of-the-art methods. In the future, we plan to expand the instance-level salient object datasets and produce more appropriate evaluation metrics for measuring the quality of the final salient instance results.

%
\section*{Acknowledgments}
This research was supported by the National Natural Science Foundation of China Grant 61902139.








%


\bibliographystyle{IEEEtran}
\bibliography{references}

\begin{thebibliography}{10}
\providecommand{\url}[1]{#1}
\csname url@samestyle\endcsname
\providecommand{\newblock}{\relax}
\providecommand{\bibinfo}[2]{#2}
\providecommand{\BIBentrySTDinterwordspacing}{\spaceskip=0pt\relax}
\providecommand{\BIBentryALTinterwordstretchfactor}{4}
\providecommand{\BIBentryALTinterwordspacing}{\spaceskip=\fontdimen2\font plus
\BIBentryALTinterwordstretchfactor\fontdimen3\font minus
  \fontdimen4\font\relax}
\providecommand{\BIBforeignlanguage}[2]{{%
\expandafter\ifx\csname l@#1\endcsname\relax
\typeout{** WARNING: IEEEtran.bst: No hyphenation pattern has been}%
\typeout{** loaded for the language `#1'. Using the pattern for}%
\typeout{** the default language instead.}%
\else
\language=\csname l@#1\endcsname
\fi
#2}}
\providecommand{\BIBdecl}{\relax}
\BIBdecl

\bibitem{itti1998model}
L.~Itti, C.~Koch, and E.~Niebur, ``A model of saliency-based visual attention
  for rapid scene analysis,'' \emph{IEEE Transactions on Pattern Analysis \&
  Machine Intelligence}, no.~11, pp. 1254--1259, 1998.

\bibitem{lei2016universal}
J.~Lei, B.~Wang, Y.~Fang, W.~Lin, P.~Le~Callet, N.~Ling, and C.~Hou, ``A
  universal framework for salient object detection,'' \emph{IEEE Transactions
  on Multimedia}, vol.~18, no.~9, pp. 1783--1795, 2016.

\bibitem{borji2014salient}
A.~Borji, M.-M. Cheng, Q.~Hou, H.~Jiang, and J.~Li, ``Salient object detection:
  A survey,'' \emph{arXiv preprint arXiv:1411.5878}, 2014.

\bibitem{cheng2010repfinder}
M.-M. Cheng, F.-L. Zhang, N.~J. Mitra, X.~Huang, and S.-M. Hu, ``Repfinder:
  finding approximately repeated scene elements for image editing,'' in
  \emph{ACM Transactions on Graphics (TOG)}, vol.~29, no.~4.\hskip 1em plus
  0.5em minus 0.4em\relax ACM, 2010, p.~83.

\bibitem{wu2014hierarchical}
C.~Wu, I.~Lenz, and A.~Saxena, ``Hierarchical semantic labeling for
  task-relevant rgb-d perception.'' in \emph{Robotics: Science and systems},
  2014.

\bibitem{gao2015database}
Y.~Gao, M.~Shi, D.~Tao, and C.~Xu, ``Database saliency for fast image
  retrieval,'' \emph{IEEE Transactions on Multimedia}, vol.~17, no.~3, pp.
  359--369, 2015.

\bibitem{hua2005generic}
X.-s. Hua, L.~Lu, H.-j. Zhang, and H.~District, ``A generic framework of user
  attention model and its application in video summarization,'' \emph{IEEE
  Transaction on multimedia}, vol.~7, no.~5, pp. 907--919, 2005.

\bibitem{sugano2010calibration}
Y.~Sugano, Y.~Matsushita, and Y.~Sato, ``Calibration-free gaze sensing using
  saliency maps,'' in \emph{2010 IEEE Computer Society Conference on Computer
  Vision and Pattern Recognition}.\hskip 1em plus 0.5em minus 0.4em\relax IEEE,
  2010, pp. 2667--2674.

\bibitem{zhao2015saliency}
R.~Zhao, W.~Ouyang, H.~Li, and X.~Wang, ``Saliency detection by multi-context
  deep learning,'' in \emph{Proceedings of the IEEE Conference on Computer
  Vision and Pattern Recognition}, 2015, pp. 1265--1274.

\bibitem{li2015visual}
G.~Li and Y.~Yu, ``Visual saliency based on multiscale deep features,'' in
  \emph{Proceedings of the IEEE conference on computer vision and pattern
  recognition}, 2015, pp. 5455--5463.

\bibitem{huo2017iterative}
S.~Huo, Y.~Zhou, J.~Lei, N.~Ling, and C.~Hou, ``Iterative feedback
  control-based salient object segmentation,'' \emph{IEEE Transactions on
  Multimedia}, vol.~20, no.~6, pp. 1350--1364, 2017.

\bibitem{ye2017salient}
L.~Ye, Z.~Liu, L.~Li, L.~Shen, C.~Bai, and Y.~Wang, ``Salient object
  segmentation via effective integration of saliency and objectness,''
  \emph{IEEE Transactions on Multimedia}, vol.~19, no.~8, pp. 1742--1756, 2017.

\bibitem{quan2017unsupervised}
R.~Quan, J.~Han, D.~Zhang, F.~Nie, X.~Qian, and X.~Li, ``Unsupervised salient
  object detection via inferring from imperfect saliency models,'' \emph{IEEE
  Transactions on Multimedia}, vol.~20, no.~5, pp. 1101--1112, 2017.

\bibitem{xiao2018deep}
H.~Xiao, J.~Feng, Y.~Wei, M.~Zhang, and S.~Yan, ``Deep salient object detection
  with dense connections and distraction diagnosis,'' \emph{IEEE Transactions
  on Multimedia}, vol.~20, no.~12, pp. 3239--3251, 2018.

\bibitem{zhang2017amulet}
P.~Zhang, D.~Wang, H.~Lu, H.~Wang, and X.~Ruan, ``Amulet: Aggregating
  multi-level convolutional features for salient object detection,'' in
  \emph{Proceedings of the IEEE International Conference on Computer Vision},
  2017, pp. 202--211.

\bibitem{liu2016dhsnet}
N.~Liu and J.~Han, ``Dhsnet: Deep hierarchical saliency network for salient
  object detection,'' in \emph{Proceedings of the IEEE Conference on Computer
  Vision and Pattern Recognition}, 2016, pp. 678--686.

\bibitem{zhang2018bi}
L.~Zhang, J.~Dai, H.~Lu, Y.~He, and G.~Wang, ``A bi-directional message passing
  model for salient object detection,'' in \emph{Proceedings of the IEEE
  Conference on Computer Vision and Pattern Recognition}, 2018, pp. 1741--1750.

\bibitem{li2017instance}
G.~Li, Y.~Xie, L.~Lin, and Y.~Yu, ``Instance-level salient object
  segmentation,'' in \emph{Proceedings of the IEEE Conference on Computer
  Vision and Pattern Recognition}, 2017, pp. 2386--2395.

\bibitem{karpathy2015deep}
A.~Karpathy and L.~Fei-Fei, ``Deep visual-semantic alignments for generating
  image descriptions,'' in \emph{Proceedings of the IEEE conference on computer
  vision and pattern recognition}, 2015, pp. 3128--3137.

\bibitem{liang2018proposal}
X.~Liang, L.~Lin, Y.~Wei, X.~Shen, J.~Yang, and S.~Yan, ``Proposal-free network
  for instance-level object segmentation,'' \emph{IEEE transactions on pattern
  analysis and machine intelligence}, vol.~40, no.~12, pp. 2978--2991, 2018.

\bibitem{fan2019s4net}
R.~Fan, M.-M. Cheng, Q.~Hou, T.-J. Mu, J.~Wang, and S.-M. Hu, ``S4net: Single
  stage salient-instance segmentation,'' in \emph{Proceedings of the IEEE
  Conference on Computer Vision and Pattern Recognition}, 2019, pp. 6103--6112.

\bibitem{cheng2014bing}
M.-M. Cheng, Z.~Zhang, W.-Y. Lin, and P.~Torr, ``Bing: Binarized normed
  gradients for objectness estimation at 300fps,'' in \emph{Proceedings of the
  IEEE conference on computer vision and pattern recognition}, 2014, pp.
  3286--3293.

\bibitem{uijlings2013selective}
J.~R. Uijlings, K.~E. Van De~Sande, T.~Gevers, and A.~W. Smeulders, ``Selective
  search for object recognition,'' \emph{International journal of computer
  vision}, vol. 104, no.~2, pp. 154--171, 2013.

\bibitem{zhang2017salient}
J.~Zhang, S.~Ma, M.~Sameki, S.~Sclaroff, M.~Betke, Z.~Lin, X.~Shen, B.~Price,
  and R.~Mech, ``Salient object subitizing,'' \emph{INTERNATIONAL JOURNAL OF
  COMPUTER VISION}, 2017.

\bibitem{ng2002spectral}
A.~Y. Ng, M.~I. Jordan, and Y.~Weiss, ``On spectral clustering: Analysis and an
  algorithm,'' in \emph{Advances in neural information processing systems},
  2002, pp. 849--856.

\bibitem{huang2017densely}
G.~Huang, Z.~Liu, L.~Van Der~Maaten, and K.~Q. Weinberger, ``Densely connected
  convolutional networks,'' in \emph{Proceedings of the IEEE conference on
  computer vision and pattern recognition}, 2017, pp. 4700--4708.

\bibitem{long2015fully}
J.~Long, E.~Shelhamer, and T.~Darrell, ``Fully convolutional networks for
  semantic segmentation,'' in \emph{Proceedings of the IEEE conference on
  computer vision and pattern recognition}, 2015, pp. 3431--3440.

\bibitem{li2016deep}
G.~Li and Y.~Yu, ``Deep contrast learning for salient object detection,'' in
  \emph{Proceedings of the IEEE Conference on Computer Vision and Pattern
  Recognition}, 2016, pp. 478--487.

\bibitem{achanta2012slic}
R.~Achanta, A.~Shaji, K.~Smith, A.~Lucchi, P.~Fua, and S.~S{\"u}sstrunk, ``Slic
  superpixels compared to state-of-the-art superpixel methods,'' \emph{IEEE
  transactions on pattern analysis and machine intelligence}, vol.~34, no.~11,
  pp. 2274--2282, 2012.

\bibitem{huang2015salicon}
X.~Huang, C.~Shen, X.~Boix, and Q.~Zhao, ``Salicon: Reducing the semantic gap
  in saliency prediction by adapting deep neural networks,'' in
  \emph{Proceedings of the IEEE International Conference on Computer Vision},
  2015, pp. 262--270.

\bibitem{borji2013state}
A.~Borji and L.~Itti, ``State-of-the-art in visual attention modeling,''
  \emph{IEEE transactions on pattern analysis and machine intelligence},
  vol.~35, no.~1, pp. 185--207, 2013.

\bibitem{liu2011learning}
T.~Liu, Z.~Yuan, J.~Sun, J.~Wang, N.~Zheng, X.~Tang, and H.-Y. Shum, ``Learning
  to detect a salient object,'' \emph{IEEE Transactions on Pattern analysis and
  machine intelligence}, vol.~33, no.~2, pp. 353--367, 2011.

\bibitem{achanta2009frequency}
R.~Achanta, S.~Hemami, F.~Estrada, and S.~S{\"u}sstrunk, ``Frequency-tuned
  salient region detection,'' in \emph{IEEE International Conference on
  Computer Vision and Pattern Recognition (CVPR 2009)}, no. CONF, 2009, pp.
  1597--1604.

\bibitem{goferman2012context}
S.~Goferman, L.~Zelnik-Manor, and A.~Tal, ``Context-aware saliency detection,''
  \emph{IEEE transactions on pattern analysis and machine intelligence},
  vol.~34, no.~10, pp. 1915--1926, 2012.

\bibitem{cheng2015global}
M.-M. Cheng, N.~J. Mitra, X.~Huang, P.~H. Torr, and S.-M. Hu, ``Global contrast
  based salient region detection,'' \emph{IEEE Transactions on Pattern Analysis
  and Machine Intelligence}, vol.~37, no.~3, pp. 569--582, 2015.

\bibitem{perazzi2012saliency}
F.~Perazzi, P.~Kr{\"a}henb{\"u}hl, Y.~Pritch, and A.~Hornung, ``Saliency
  filters: Contrast based filtering for salient region detection,'' in
  \emph{2012 IEEE conference on computer vision and pattern recognition}.\hskip
  1em plus 0.5em minus 0.4em\relax IEEE, 2012, pp. 733--740.

\bibitem{zhang2013saliency}
J.~Zhang and S.~Sclaroff, ``Saliency detection: A boolean map approach,'' in
  \emph{Proceedings of the IEEE international conference on computer vision},
  2013, pp. 153--160.

\bibitem{kruthiventi2016saliency}
S.~S. Kruthiventi, V.~Gudisa, J.~H. Dholakiya, and R.~Venkatesh~Babu,
  ``Saliency unified: A deep architecture for simultaneous eye fixation
  prediction and salient object segmentation,'' in \emph{Proceedings of the
  IEEE Conference on Computer Vision and Pattern Recognition}, 2016, pp.
  5781--5790.

\bibitem{li2016deepsaliency}
X.~Li, L.~Zhao, L.~Wei, M.-H. Yang, F.~Wu, Y.~Zhuang, H.~Ling, and J.~Wang,
  ``Deepsaliency: Multi-task deep neural network model for salient object
  detection,'' \emph{IEEE Transactions on Image Processing}, vol.~25, no.~8,
  pp. 3919--3930, 2016.

\bibitem{hariharan2014simultaneous}
B.~Hariharan, P.~Arbel{\'a}ez, R.~Girshick, and J.~Malik, ``Simultaneous
  detection and segmentation,'' in \emph{European Conference on Computer
  Vision}.\hskip 1em plus 0.5em minus 0.4em\relax Springer, 2014, pp. 297--312.

\bibitem{dai2016instance}
J.~Dai, K.~He, and J.~Sun, ``Instance-aware semantic segmentation via
  multi-task network cascades,'' in \emph{Proceedings of the IEEE Conference on
  Computer Vision and Pattern Recognition}, 2016, pp. 3150--3158.

\bibitem{dai2016instancee}
J.~Dai, K.~He, Y.~Li, S.~Ren, and J.~Sun, ``Instance-sensitive fully
  convolutional networks,'' in \emph{European Conference on Computer
  Vision}.\hskip 1em plus 0.5em minus 0.4em\relax Springer, 2016, pp. 534--549.

\bibitem{he2017mask}
K.~He, G.~Gkioxari, P.~Doll{\'a}r, and R.~Girshick, ``Mask r-cnn,'' in
  \emph{Proceedings of the IEEE international conference on computer vision},
  2017, pp. 2961--2969.

\bibitem{girshick2015fast}
R.~Girshick, ``Fast r-cnn,'' in \emph{Proceedings of the IEEE international
  conference on computer vision}, 2015, pp. 1440--1448.

\bibitem{zhang2016unconstrained}
J.~Zhang, S.~Sclaroff, Z.~Lin, X.~Shen, B.~Price, and R.~Mech, ``Unconstrained
  salient object detection via proposal subset optimization,'' in
  \emph{Proceedings of the IEEE conference on computer vision and pattern
  recognition}, 2016, pp. 5733--5742.

\bibitem{treisman1980feature}
A.~M. Treisman and G.~Gelade, ``A feature-integration theory of attention,''
  \emph{Cognitive psychology}, vol.~12, no.~1, pp. 97--136, 1980.

\bibitem{krahenbuhl2011efficient}
P.~Kr{\"a}henb{\"u}hl and V.~Koltun, ``Efficient inference in fully connected
  crfs with gaussian edge potentials,'' in \emph{Advances in neural information
  processing systems}, 2011, pp. 109--117.

\bibitem{alexe2010object}
B.~Alexe, T.~Deselaers, and V.~Ferrari, ``What is an object?'' in \emph{2010
  IEEE Computer Society Conference on Computer Vision and Pattern
  Recognition}.\hskip 1em plus 0.5em minus 0.4em\relax IEEE, 2010, pp. 73--80.

\bibitem{he2017delving}
S.~He, J.~Jiao, X.~Zhang, G.~Han, and R.~W. Lau, ``Delving into salient object
  subitizing and detection,'' in \emph{Proceedings of the IEEE International
  Conference on Computer Vision}, 2017, pp. 1059--1067.

\bibitem{krizhevsky2012imagenet}
A.~Krizhevsky, I.~Sutskever, and G.~E. Hinton, ``Imagenet classification with
  deep convolutional neural networks,'' in \emph{Advances in neural information
  processing systems}, 2012, pp. 1097--1105.

\bibitem{hu2018squeeze}
J.~Hu, L.~Shen, and G.~Sun, ``Squeeze-and-excitation networks,'' in
  \emph{Proceedings of the IEEE conference on computer vision and pattern
  recognition}, 2018, pp. 7132--7141.

\bibitem{ioffe2015batch}
S.~Ioffe and C.~Szegedy, ``Batch normalization: Accelerating deep network
  training by reducing internal covariate shift,'' \emph{arXiv preprint
  arXiv:1502.03167}, 2015.

\bibitem{glorot2011deep}
X.~Glorot, A.~Bordes, and Y.~Bengio, ``Deep sparse rectifier neural networks,''
  in \emph{Proceedings of the fourteenth international conference on artificial
  intelligence and statistics}, 2011, pp. 315--323.

\bibitem{ronneberger2015u}
O.~Ronneberger, P.~Fischer, and T.~Brox, ``U-net: Convolutional networks for
  biomedical image segmentation,'' in \emph{International Conference on Medical
  image computing and computer-assisted intervention}.\hskip 1em plus 0.5em
  minus 0.4em\relax Springer, 2015, pp. 234--241.

\bibitem{chen2014semantic}
L.-C. Chen, G.~Papandreou, I.~Kokkinos, K.~Murphy, and A.~L. Yuille, ``Semantic
  image segmentation with deep convolutional nets and fully connected crfs,''
  \emph{arXiv preprint arXiv:1412.7062}, 2014.

\bibitem{zelnik2005self}
L.~Zelnik-Manor and P.~Perona, ``Self-tuning spectral clustering,'' in
  \emph{Advances in neural information processing systems}, 2005, pp.
  1601--1608.

\bibitem{yang2013saliency}
C.~Yang, L.~Zhang, H.~Lu, X.~Ruan, and M.-H. Yang, ``Saliency detection via
  graph-based manifold ranking,'' in \emph{Proceedings of the IEEE conference
  on computer vision and pattern recognition}, 2013, pp. 3166--3173.

\bibitem{lecun1989backpropagation}
Y.~LeCun, B.~Boser, J.~S. Denker, D.~Henderson, R.~E. Howard, W.~Hubbard, and
  L.~D. Jackel, ``Backpropagation applied to handwritten zip code
  recognition,'' \emph{Neural computation}, vol.~1, no.~4, pp. 541--551, 1989.

\bibitem{yan2013hierarchical}
Q.~Yan, L.~Xu, J.~Shi, and J.~Jia, ``Hierarchical saliency detection,'' in
  \emph{Proceedings of the IEEE Conference on Computer Vision and Pattern
  Recognition}, 2013, pp. 1155--1162.

\bibitem{movahedi2010design}
V.~Movahedi and J.~H. Elder, ``Design and perceptual validation of performance
  measures for salient object segmentation,'' in \emph{2010 IEEE Computer
  Society Conference on Computer Vision and Pattern
  Recognition-Workshops}.\hskip 1em plus 0.5em minus 0.4em\relax IEEE, 2010,
  pp. 49--56.

\bibitem{lin2018saliency}
X.~Lin, Z.-J. Wang, L.~Ma, and X.~Wu, ``Saliency detection via multi-scale
  global cues,'' \emph{IEEE Transactions on Multimedia}, 2018.

\bibitem{borji2015salient}
A.~Borji, M.-M. Cheng, H.~Jiang, and J.~Li, ``Salient object detection: A
  benchmark,'' \emph{IEEE transactions on image processing}, vol.~24, no.~12,
  pp. 5706--5722, 2015.

\bibitem{fan2017structure}
D.-P. Fan, M.-M. Cheng, Y.~Liu, T.~Li, and A.~Borji, ``{Structure-measure: A
  New Way to Evaluate Foreground Maps},'' in \emph{IEEE International
  Conference on Computer Vision (ICCV)}.\hskip 1em plus 0.5em minus 0.4em\relax
  IEEE, 2017, pp. 4548--4557, \url{http://dpfan.net/smeasure/}.

\bibitem{Fan2018Enhanced}
D.-P. Fan, C.~Gong, Y.~Cao, B.~Ren, M.-M. Cheng, and A.~Borji,
  ``{Enhanced-alignment Measure for Binary Foreground Map Evaluation},'' in
  \emph{International Joint Conference on Artificial Intelligence (IJCAI)},
  2018, pp. 698--704, \url{http://dpfan.net/e-measure/}.

\bibitem{cheng2013efficient}
M.-M. Cheng, J.~Warrell, W.-Y. Lin, S.~Zheng, V.~Vineet, and N.~Crook,
  ``Efficient salient region detection with soft image abstraction,'' in
  \emph{Proceedings of the IEEE International Conference on Computer vision},
  2013, pp. 1529--1536.

\bibitem{jiang2013salient}
H.~Jiang, J.~Wang, Z.~Yuan, Y.~Wu, N.~Zheng, and S.~Li, ``Salient object
  detection: A discriminative regional feature integration approach,'' in
  \emph{Proceedings of the IEEE conference on computer vision and pattern
  recognition}, 2013, pp. 2083--2090.

\bibitem{hou2017deeply}
Q.~Hou, M.-M. Cheng, X.~Hu, A.~Borji, Z.~Tu, and P.~H. Torr, ``Deeply
  supervised salient object detection with short connections,'' in
  \emph{Proceedings of the IEEE Conference on Computer Vision and Pattern
  Recognition}, 2017, pp. 3203--3212.

\bibitem{zhang2018progressive}
X.~Zhang, T.~Wang, J.~Qi, H.~Lu, and G.~Wang, ``Progressive attention guided
  recurrent network for salient object detection,'' in \emph{Proceedings of the
  IEEE Conference on Computer Vision and Pattern Recognition}, 2018, pp.
  714--722.

\bibitem{deng2018r3net}
Z.~Deng, X.~Hu, L.~Zhu, X.~Xu, J.~Qin, G.~Han, and P.-A. Heng, ``R3net:
  Recurrent residual refinement network for saliency detection,'' in
  \emph{Proceedings of the 27th International Joint Conference on Artificial
  Intelligence}.\hskip 1em plus 0.5em minus 0.4em\relax AAAI Press, 2018, pp.
  684--690.

\end{thebibliography}
%








\end{document}